\documentclass{article}

\usepackage{microtype}
\usepackage{graphicx}
\usepackage{subfigure}
\usepackage{booktabs} 
\usepackage{placeins} 
\usepackage{hyperref}

\usepackage[arxiv]{icml2025}

\usepackage{amsmath}
\usepackage{amssymb}
\usepackage{mathtools}
\usepackage{amsthm}
\usepackage{multirow}
\usepackage[capitalize,noabbrev]{cleveref}

\theoremstyle{plain}

\theoremstyle{definition}

\theoremstyle{remark}

\usepackage{tcolorbox}

\makeatletter
\def\section{\@startsection{section}{1}{\z@}{-0.1in}{0.01in}{\large\bf\raggedright}}  

\def\subsection{\@startsection{subsection}{2}{\z@}{-0.08in}{0.01in}{\normalsize\bf\raggedright}}
\def\paragraph{\@startsection{paragraph}{4}{\z@}{0.5ex plus 0.5ex minus .2ex}{-1em}{\normalsize\bf}}
\long\def\@makefigurecaption#1#2{
 \vskip 5pt                            
        \baselineskip 11pt
        \setbox\@tempboxa\hbox{#1. #2}
        \ifdim \wd\@tempboxa >\hsize
        \sbox{\newcaptionbox}{\small\sl #1.~}
        \newcaptionboxwid=\wd\newcaptionbox
        \usebox\newcaptionbox {\footnotesize #2}\vskip -10pt  
        \else
          \centerline{{\small\sl #1.} {\small #2}}\vskip -10pt  
        \fi}
\let\@makefigurecaption\@makefigurecaption
\makeatother

\icmltitlerunning{Reflective Planning: Vision-Language Models for Multi-Stage Long-Horizon Robotic Manipulation}

\begin{document}

\twocolumn[
\icmltitle{Reflective Planning: Vision-Language Models for \\Multi-Stage Long-Horizon Robotic Manipulation}

\icmlsetsymbol{advisor}{$\dagger$}
\begin{icmlauthorlist}
\icmlauthor{Yunhai Feng}{cornell}
\icmlauthor{Jiaming Han}{cuhk}
\icmlauthor{Zhuoran Yang}{yale}
\icmlauthor{Xiangyu Yue}{cuhk}
\icmlauthor{Sergey Levine}{ucb}
\icmlauthor{Jianlan Luo}{ucb,advisor}
\end{icmlauthorlist}

\icmlaffiliation{cornell}{Cornell University}
\icmlaffiliation{cuhk}{The Chinese University of Hong Kong}
\icmlaffiliation{yale}{Yale University}
\icmlaffiliation{ucb}{University of California, Berkeley}

\icmlcorrespondingauthor{Yunhai Feng}{yunhaif@cs.cornell.edu}
\icmlcorrespondingauthor{Xiangyu Yue}{xyyue@ie.cuhk.edu.hk}
\icmlcorrespondingauthor{Jianlan Luo}{jianlanluo@eecs.berkeley.edu}

\icmlkeywords{Machine Learning, ICML}

\vskip 0.3in
]

\printAffiliationsAndNotice{\icmlAdvisor} 

\begin{abstract}
Solving complex long-horizon robotic manipulation problems requires sophisticated high-level planning capabilities, the ability to reason about the physical world, and reactively choose appropriate motor skills. Vision-language models (VLMs) pretrained on Internet data could in principle offer a framework for tackling such problems. However, in their current form, VLMs lack both the nuanced understanding of intricate physics required for robotic manipulation and the ability to reason over long horizons to address error compounding issues.
In this paper, we introduce a novel test-time computation framework that enhances VLMs' physical reasoning capabilities for multi-stage manipulation tasks. At its core, our approach iteratively improves a pretrained VLM with a ``reflection'' mechanism - it uses a generative model to imagine future world states, leverages these predictions to guide action selection, and critically reflects on potential suboptimalities to refine its reasoning. Experimental results demonstrate that our method significantly outperforms several state-of-the-art commercial VLMs as well as other post-training approaches such as Monte Carlo Tree Search (MCTS). Videos are available at \url{https://reflect-vlm.github.io}.
\end{abstract}

\section{Introduction}\label{sec:intro}

\begin{figure}[t!]
    \centering
    \includegraphics[width=0.45\textwidth]{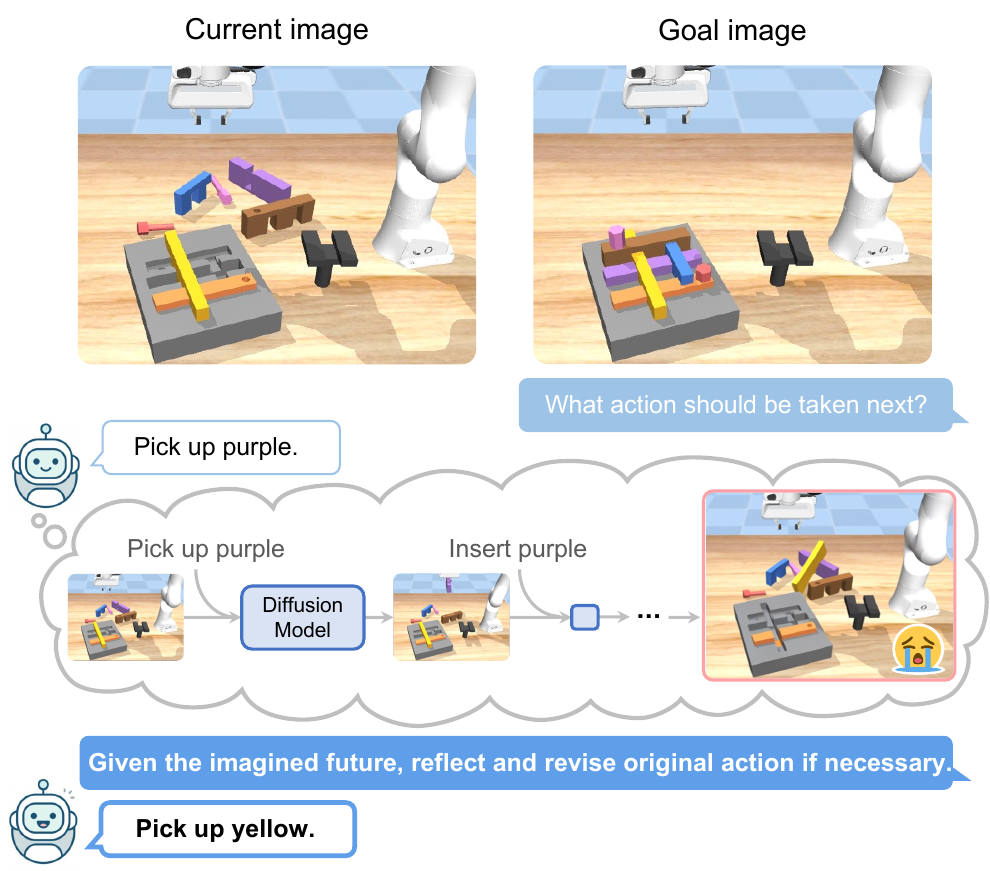}
    \caption{\textbf{Reflective planning.} Our method uses a VLM to propose actions and a diffusion dynamics model to imagine the future state of executing the plan. The imagined future helps the VLM reflect the initial plan and propose better action.}
    \label{fig:teaser}
\end{figure}

Complex multi-stage manipulation tasks remain a fundamental challenge in robotics~\citep{luo2024cable, kroemer2020reviewgmanipulation, Cui2021TowardNL}, particularly when they require reasoning about sophisticated physical interactions and their consequences over long time horizons. These tasks often involve intricate sequences of actions where each step must account for physical constraints and potential consequences, making them particularly challenging for planning systems. Success requires not only understanding the immediate effects of actions but also their long-term implications, the ability to adapt plans based on execution outcomes, and generalizing to novel scenarios.

While classical planning approaches, such as task and motion planning (TAMP)~\citep{kaelbling2011tamp, garrett2020integratedtamp}, can in principle address such problems, their reliance on predefined symbolic representations and explicit state estimation makes them difficult to apply in settings without known models that require visual perception~\citep{driess2020deepvisualreasoninglearning, wang2021learningcompositional}. This limitation has motivated the search for more flexible approaches to robotic planning. Recent advances in vision-language models (VLMs) have shown remarkable capabilities in processing visual scenes and natural language instructions by leveraging internet-scale knowledge~\citep{pali2023, qwen2023, openai2024gpt4ocard, google2024gemini, llava}. These models can effectively parse complex visual environments and comprehend high-level task descriptions expressed in natural language, making them promising candidates for robotic planning problems~\citep{palm-e2023, rt1, rt2,shi2024yal, liu2024moka},. However, state-of-the-art VLMs still struggle with complex physical reasoning tasks, and this limitation becomes particularly pronounced when precise physics concepts and long-horizon planning are involved~\citep{gao2024physicsVLM, chen2024spatialvlm}.

In this paper, we study how to effectively leverage VLMs' Internet-scale knowledge while addressing their limitations in physical reasoning and long-horizon planning. We focus on a challenging class of robotic manipulation problems that involve sequentially manipulating interlocking objects to achieve desired configurations, as illustrated in Fig.~\ref{fig:tasks}. These tasks are particularly difficult as they require precise understanding of physical constraints, careful reasoning about action sequences, and the ability to plan over extended horizons while maintaining physical feasibility at each step.

To address these challenges, we present a novel test-time computation framework that significantly enhances VLMs' capabilities for multi-stage robotic manipulation tasks. 
The key insight of our method, ReflectVLM, is that by combining VLMs with a reflection mechanism and targeted post-training, we can create a system that better understands physical constraints and their implications for action planning. We use the term ``reflection'' to refer to a process where a VLM iteratively refines its decisions by critically examining the predicted outcomes of its proposed actions, akin to self-critique methods in large language models~\citep{huang2024llmcorrect, wang2023selfinstructaligninglanguagemodels, madaan2024self}. Our approach introduces two key components: (1) a look-ahead mechanism that uses a diffusion-based dynamics model to generate visual predictions of future states resulting from planned actions, and (2)a reflection process that allows the VLM to critique and refine its planned actions by analyzing these predicted outcomes. This combination of visual prediction and iterative refinement allows the VLM to develop a more sophisticated understanding of physical constraints and improve its decision-making capabilities without requiring extensive retraining.

Experimental results demonstrate that our approach significantly outperforms both the latest commercial state-of-the-art VLM models and traditional planning approaches like Monte Carlo Tree Search (MCTS) on this class of problems. Notably, our method achieves superior performance compared to post-training techniques such as supervised fine-tuning (SFT) while using the same amount of labeled data and maintaining computational efficiency. The success of our approach suggests that enhancing VLMs with structured reasoning mechanisms at test time can be a powerful strategy for improving their performance on physically-grounded tasks.

Our primary contribution is the mentioned test-time computation framework that enhances VLMs' physical reasoning capabilities for multi-stage manipulation tasks. Through extensive experiments, we demonstrate that our approach not only outperforms existing methods but also maintains computational efficiency. Importantly, while we demonstrate our framework's effectiveness on manipulation tasks, it is designed to be general and can be readily extended to other domains requiring visual understanding and sequential decision-making. This generality suggests broader applications in robotics and autonomous systems where physical reasoning and long-horizon planning are essential.

\begin{figure*}[t!]
    \centering
    \includegraphics[width=0.99\textwidth]{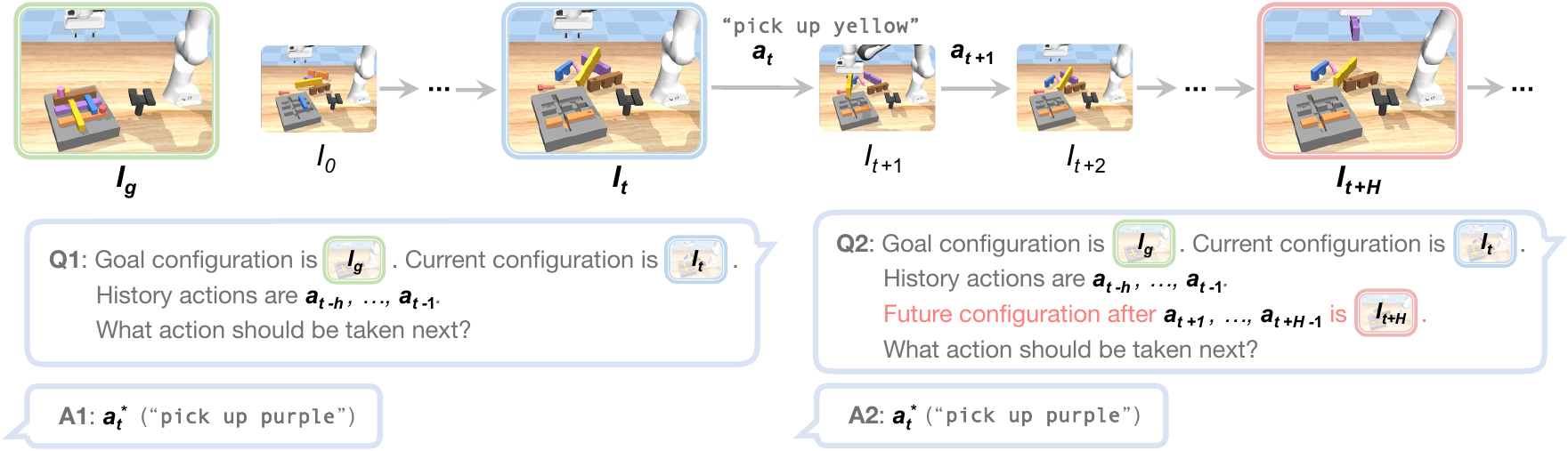}
    \caption{\textbf{Training data generation.} Training data for the reflection mechanism is collected by relabeling the rollouts. For each timestep, two training examples are generated: (Q1, A1) for action proposal and (Q2, A2) for reflection. $H$ is the imagination horizon, and $h$ is the history length. $a_t^*$ is the action label given by the expert policy.}

    \label{fig:data_dollection}
\end{figure*}

\section{Related Work}\label{sec:related_work}
Our framework incorporates a VLM with the reflection mechanism to solve long-horizon robotic planning problems. We therefore survey reflection techniques in the broader context in large models, VLM for robotic planning, as well as existing techniques for solving robot task and motion planning.

\subsection{Reflection}
Recent work has shown that large language models can benefit from reflection mechanisms - processes where models iteratively refine their outputs through self-critique and revision~\citep{renze2024self, shinn2024reflexion, pan2023automatically, madaan2024self, asai2023self, wang2023selfinstructaligninglanguagemodels, huang2024llmcorrect}. For example,~\citet{madaan2024self} introduced an iterative refinement approach where models critique and improve their own outputs through self-feedback. Chain-of-thought prompting and its variants~\citep{wei2022chain, wang2022self, yao2024tree} demonstrated that guiding models to show their reasoning process leads to better performance. Similarly,~\citet{cheng2024vlmimproving, yu2025exactteachingaiagents} extended such reflection mechanisms to vision-language models.

However, these approaches focus primarily on language-only or visual comprehension tasks, without addressing physical reasoning or robotics applications. Our work extends reflection to long-horizon robotic planning by incorporating a diffusion model that generates imagined future visual states. This allows the VLM to reflect on and revise its plans based on concrete visual predictions rather than relying solely on symbolic reasoning.

\subsection{VLM for Robotic Planning} 
In robotics, several recent works have explored using VLMs for planning~\citep{palm-e2023, rt1, rt2, hu2023lookleapunveilingpower, huang2023voxposer, belkhale2024rth, nasiriany2024pivot, liu2024moka, shi2024yal, wake2024gpt}. However, these approaches either rely on symbolic state representations or make decisions in a single-step manner based only on current observations, without explicitly reasoning about future consequences or utilizing reflection mechanisms.

While ReplanVLM~\citep{mei2024replanvlm} and GameVLM~\citep{mei2024gamevlm} use VLMs to replan robot actions based on execution feedback, they still rely on symbolic state representations rather than visual imagination of future states.~\citet{black2023susie} utilized a diffusion model to generate future visual states and executed them with a low-level goal-conditioned policy, but did not leverage these predictions for plan reflection or revision.~\citet{du2023videolanguageplanning} combines a VLM with video prediction for beam search, but suffers from prediction error accumulation and struggles with physics-based reasoning tasks.

Our framework addresses these limitations by enabling VLMs to imagine and evaluate potential future states through a diffusion-based dynamics model. This allows for sophisticated multi-step planning while maintaining the benefits of VLMs' pre-trained visual-language understanding. The reflection mechanism further enables the VLM to critique and refine its plans based on these imagined futures, leading to more robust long-horizon manipulation.

\subsection{Robotic Task and Motion Planning}
Robotic Task and Motion Planning (TAMP) has been extensively studied~\citep{kaelbling2011tamp,garrett2020integratedtamp, garrett2020pddlstream}. Traditional approaches often combine symbolic planning with motion planning but struggle with real-world physical interactions and visual inputs. Learning-based methods~\citep{wang2021learningcompositional,driess2020deepvisualreasoninglearning} show promise in handling uncertainty and complex dynamics but typically require significant task-specific engineering.

Our approach bridges this gap by leveraging VLMs' broad knowledge while adding structured physical reasoning through visual imagination and reflection. This enables robust long-horizon planning without requiring extensive task-specific engineering or large amounts of training data.
\section{Preliminaries and Problem Statement}\label{sec:problem}
We formulate the multi-stage robotic manipulation planning problem as a partially observable Markov decision process (POMDP), defined by the tuple $(\mathcal{S}, \mathcal{A}, \mathcal{T}, \mathcal{O}, \mathcal{Z})$. 
Here, $\mathcal{S}$ is the state space containing the full physical state of the environment, including object poses and physical properties; $\mathcal{A}$ is the action space consisting of high-level manipulation primitives $\{{\tt pick\ up}, {\tt insert}, {\tt reorient}, {\tt put\ down}\} \times \{\text{objects}\}$, assuming a failure rate $\epsilon$ for each primitive; $\mathcal{T}(s_{t+1}|s_t,a_t)$ represents the transition dynamics capturing physical interactions; $\mathcal{O}$ is the observation space of RGB images; and $\mathcal{Z}(o_t|s_t)$ is the observation model mapping states to images.

Given a goal state $s_g$, the objective is to find a policy $\pi$ that generates a sequence of actions to reach $s_g$. Due to partial observability, the policy only has access to image observations, taking the form $\pi(a_t|I_t,I_g)$ where $I_t$ is the current observation and $I_g$ is the goal image. The policy is instantiated as a VLM agent $\pi_\text{VLM}$, which takes a multi-modal input of images and text, and generates action primitives in the form of text.

Our framework includes a pre-training phase and a post-training phase. The post-training phase builds on the framework of interactive imitation learning~\citep{dagger, Kelly2018HGDAgger}, which learns a policy by interacting with environment and receiving expert supervision in real-time. Thus under the standard assumption, we assume access to an interactive expert policy $\pi_E$ that generates near-optimal actions $a^*=\pi_E(s)$ for any state $s$ at training time. 
In this paper, we instantiated such an expert policy with access to the full state of the environment to generate optimal actions, though it could be obtained via other formats as well, e.g., human demonstrations.
However, the VLM policy will only have access to image observations.

\section{Reflective Planning with Vision Language Models}\label{sec:methods}
To address the challenges of physical interaction and long-horizon reasoning, we present a framework that incorporates VLMs with reflective planning. Our approach combines two key components: (1) a diffusion-based dynamics model that enables the VLM to imagine and evaluate future states, and (2) an interactive learning mechanism that allows the VLM to reflect on and revise its decisions based on these imagined outcomes. As shown in Fig.~\ref{fig:teaser}, these components work together to enable more robust manipulation planning while preserving the benefits of pre-trained VLMs.

\subsection{Interactive VLM Policy Post-Training}\label{sec:policy_training}
While VLMs can generate actions based on visual inputs, they may hallucinate physically implausible solutions without actual interaction experience. 
To overcome this limitation and enable long-horizon reasoning, we introduce an interactive learning algorithm that teaches the VLM to reflect on and improve its decisions through direct interaction with the physical environment. This process further enhances a base VLM policy, which is initially trained on a fixed set of expert demonstrations.
Similar to DAgger~\citep{dagger}, we iteratively collect new data by rolling out the VLM policy in the environment and finetune the VLM policy with the aggregated data. As formulated in Algorithm~\ref{alg:training}, $N$ trajectories are collected in each iteration. At each timestep, we generate a learner action $a_t^\dag$ by prompting the VLM with the images of the goal and current states, as well as an expert action $a_t^*$ from the oracle policy. The pairs $((I_g, I_t), a_t^*)$ are then added to the dataset for finetuning. To facilitate convergence, we execute the learner action $a_t^\dag$ with a probability of $p$ and the expert action $a_t^*$ with a probability of $1-p$, instead of always following the actions from the learner. 

\begin{algorithm}
\caption{Interactive VLM Post-Training}
\label{alg:training}
\begin{algorithmic}[1]
\REQUIRE {initial state distribution $\rho_0$, goal state distribution $\rho_g$, numbef of iterations $K$, number of trajectories per iteration $N$, episode length $T$, imagination horizon $H$, expert policy $\pi_E$, expert demonstrations $\mathcal{D}^*$}
\STATE train base policy $\pi_\text{VLM}$ on $\mathcal{D}^*$
\STATE $\mathcal{D}\gets\mathcal{D}^*$
\FOR {$i \gets 1$ to $K$}
    \STATE $\mathcal{D}_i\gets\emptyset$
    \STATE {// rollout out policy $\pi_\text{VLM}$ to collect data $\mathcal{D}_{i}$}
    \FOR {$n \gets 1$ to $N$}
        \STATE $s_0\sim\rho_0$; $I_0\gets\mathcal{Z}(s_0)$
        \STATE $s_g\sim\rho_g$; $I_g\gets\mathcal{Z}(s_g)$
        \FOR {$t \gets 0$ to $T-1$}
            \STATE $a_t^{\dag}\sim\pi_\text{VLM}(I_g, I_t)$; $a_t^* \sim \pi_E(s_g, s_t)$
            \STATE $a_t\gets a_t^{\dag}$ \textbf{if} ${\tt random()} <p$ \textbf{else} $a_t^*$
            \STATE $s_{t+1}\gets\mathcal{T}(s_t, a_t)$; $I_{t+1}\gets\mathcal{Z}(s_{t+1})$
        \ENDFOR
        \STATE $\mathcal{D}_i\!\gets\!\mathcal{D}_i\cup\{((I_g, I_t), a_t^*)\}_{0\le t<T}$
        \STATE {\textcolor{red}{$\mathcal{D}_i\!\gets\!\mathcal{D}_i\cup\{((I_g, I_t, I_{t+H}, a_{t:t+H-1}), a_t^*)\}_{0\le t<T}$}}
    \ENDFOR
    \STATE {$\mathcal{D}\gets \mathcal{D}\cup\mathcal{D}_{i}$}
    \STATE {finetune $\pi_\text{VLM}$ on $\mathcal{D}$}
\ENDFOR
\end{algorithmic}
\end{algorithm}

To generate training data for reflection, we can simply relabel a trajectory after it is terminated, as also illustrated in Fig.~\ref{fig:data_dollection}. Specifically, the image $I_{t+H}$, which is a future observation following the action sequence $a_{t:t+H-1}$, is added to the context for reflection at timestep $t$, and the VLM is still supervised to output the same expert action $a_t^*$. Intuitively, this image provides additional information about the effect of executing the action sequence as a feedback, which can be leveraged by the VLM to decide whether the initially proposed action sequence leads to a promising future state. 

In essence, we are generating two forms of question answering examples from interaction with the environment. The first is to predict an optimal action given images of the goal and current state, and the second is to reflect and revise an initial action sequence proposal by looking into an additional future image.
Since a VLM can flexibly take any text and images as input, these two tasks can be handled by a single VLM with two different prompt templates, as summarized in Fig.~\ref{fig:data_dollection}. See App.~\ref{sec:app_prompts} for full prompts, and App.~\ref{subsec:vlm_policy} for detailed VLM architecture.

The VLM is trained to generate actions aligned with expert actions in the dataset with a cross entropy loss:
\begin{equation}
\begin{split}
    \min_{\pi_\text{VLM}}\ & \mathbb{E}_{\mathcal{D}}\Big[\mathcal{L}_\text{CE}(\pi_\text{VLM}^\text{propose}(a_t|I_g, I_t), a_t^*) \\
   +\ & \mathcal{L}_\text{CE}(\pi_\text{VLM}^\text{reflect}(a_t|I_g, I_t, I_{t+H}, a_{t:t+H-1}), a_t^*)\Big]. 
\end{split}
\end{equation}

\subsection{Diffusion Dynamics Model}\label{sec:diffusion_model}
A key component in reflective planning is predicting future states accurately when evaluating potential action sequences. While our interactive learning mechanism enables the VLM to learn from physical interactions, we need an additional capability during inference - the ability to imagine and evaluate hypothetical futures without actually executing actions in the environment. To address this, we develop a diffusion-based dynamics model (DDM) that efficiently generates predicted visual observations by conditioning on the current observation and a proposed action sequence. This allows the VLM to simulate the consequences of its actions before committing to them.

\begin{figure}[h]
    \centering
    \includegraphics[width=0.48\textwidth]{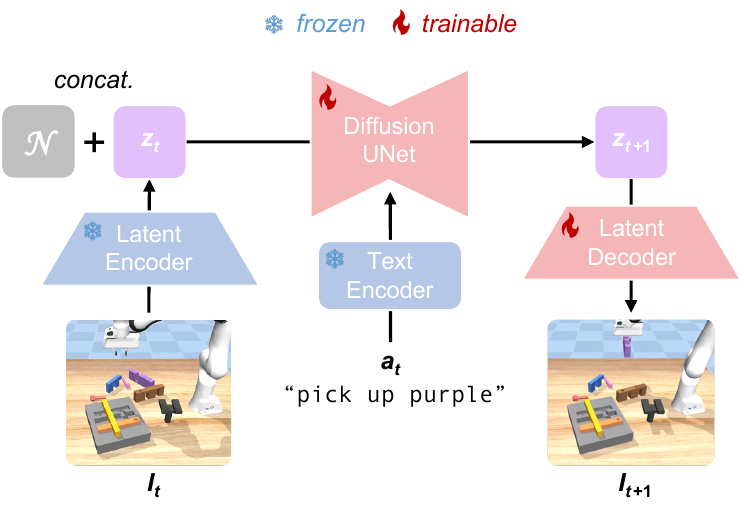}
    \vspace{-0.2in}
    \caption{\textbf{Architecture of Diffusion Dynamics Model,} which consists of a latent encoder, text encoder, Diffusion UNet and latent decoder. The latent encoder and text encoder are frozen during training, while Diffusion UNet and latent decoder are finetuned on our task data. $\mathcal{N}$: random noise.}
    \label{fig:diffusion_model}
\end{figure}

Building on advances in diffusion-based generative models~\citep{rombach2021highresolution, ho2020ddpm, song2021score}, we formulate the forward dynamics prediction as an image-to-image translation task. Our diffusion dynamics model takes the current observation $I_t$ and action $a_t$ as input to predict the next observation $I_{t+1}$. Rather than training a diffusion model from scratch, which would require substantial computational resources and training data, we leverage the pretrained Instructpix2pix model~\citep{brooks2022instructpix2pix} that has been trained on large-scale image editing datasets as our base model.

\noindent \textbf{Data.} We curate a dataset for training the diffusion model. To encourage broader coverage of visisted states, the data collection policy is a noised version of the oracle policy.
Due to the difficulty of this task, we also include a few test data points to improve the fidelity and accuracy of the DDM.
Details can be found in App.~\ref{app:ddm}.

\noindent \textbf{Architecture.} The model architecture is shown in Fig.~\ref{fig:diffusion_model}. For the input $(I_t, a_t)$, we first encode them into latent representation $z_t$ and $z_{a_t}$ with pretrained latent encoder and text encoder. Then we feed $z_t$, a sampled noise $\mathcal{N}$ and the action condition $z_{a_t}$ into the diffusion UNet for de-noising. Finally, we decode the predicted $z_{t+1}$ into a future observation $I_{t+1}$ with a latent decoder.

\noindent \textbf{Training.} The training of DDM consists of two separate phases: UNet training and decoder training. The UNet training phase is to learn transformations from $z_t$ to $z_{t+1}$ conditioned on $z_{a_t}$, while the latent decoder training is to adapt the pretrained VAE models into our task domain because our task requires precise reconstruction of small pieces on the table. Since we keep the latent encoder frozen, we can train the two phases in parallel.

\subsection{Reflective Planning}\label{sec:reflection}

With the VLM policy trained via interactive learning and the diffusion model serving as a dynamics proxy to imagine future outcomes, we now introduce our reflective planning mechanism for decision making at inference time. Alg.~\ref{alg:inference} shows the detailed process. We use $\tilde{I}$ and ${\tilde{a}}$ to denote the generated image and action, which are not actually observed or executed in the environment. To get the future image after $H$ steps, where $H$ is the planning horizon, we perform $H$ iterations of action proposal and diffusion generation. At each iteration, the VLM policy is prompted by the goal image $I_g$ and the generated image $\tilde{I}_{t+k}$ at the previous iteration to propose an action $\tilde{a}_{t+k}$. The diffusion model $\tilde{\mathcal{T}}$ then generates the future image $\tilde{I}_{t+k+1}$ conditioned on the previous image $\tilde{I}_{t+k}$ and the action $\tilde{a}_{t+k}$. For the first iteration, the input image $\tilde{I}_t$ is just the current observation $I_t$. After this process of imagination, the generated future image $\tilde{I}_{t+H}$ and the plan $\tilde{a}_{t:t+H-1}$ are concatenated with the goal and current observation, and fed into the VLM policy for reflection. The VLM policy will then output the final action $a_t$ to be executed. Again, action proposal and reflection are performed by the same VLM policy with two different prompt templates, as indicated by the superscripts ``propose'' and ``reflect''.

\begin{algorithm}
    \caption{Reflective Planning (Inference)}
    \label{alg:inference}
    \begin{algorithmic}[1]
    \REQUIRE {current image $I_t$, goal image $I_g$, imagination horizon $H$}
    \STATE $\tilde{I}_t\gets I_t$
    \FOR {$k \gets 0$ to $H-1$}
        \STATE {$\tilde{a}_{t+k} \gets \pi_\text{VLM}^\text{propose}(I_g, \tilde{I}_{t+k})$}
        \STATE {$\tilde{I}_{t+k+1} \gets \mathcal{\tilde{T}}(\tilde{I}_{t+k}, \tilde{a}_{t+k})$}
    \ENDFOR
    \STATE{$a_t \gets \pi_\text{VLM}^\text{reflect}(I_g, I_t, \tilde{I}_{t+H}, \tilde{a}_{t:t+H-1})$}
    \STATE \textbf{Output:} {$a_t$}
    \end{algorithmic}
\end{algorithm}

\section{Multi-Stage Robotic Manipulation Planning Tasks}\label{sec:tasks}

\begin{figure*}[t!]
    \centering
    \includegraphics[width=0.99\textwidth]{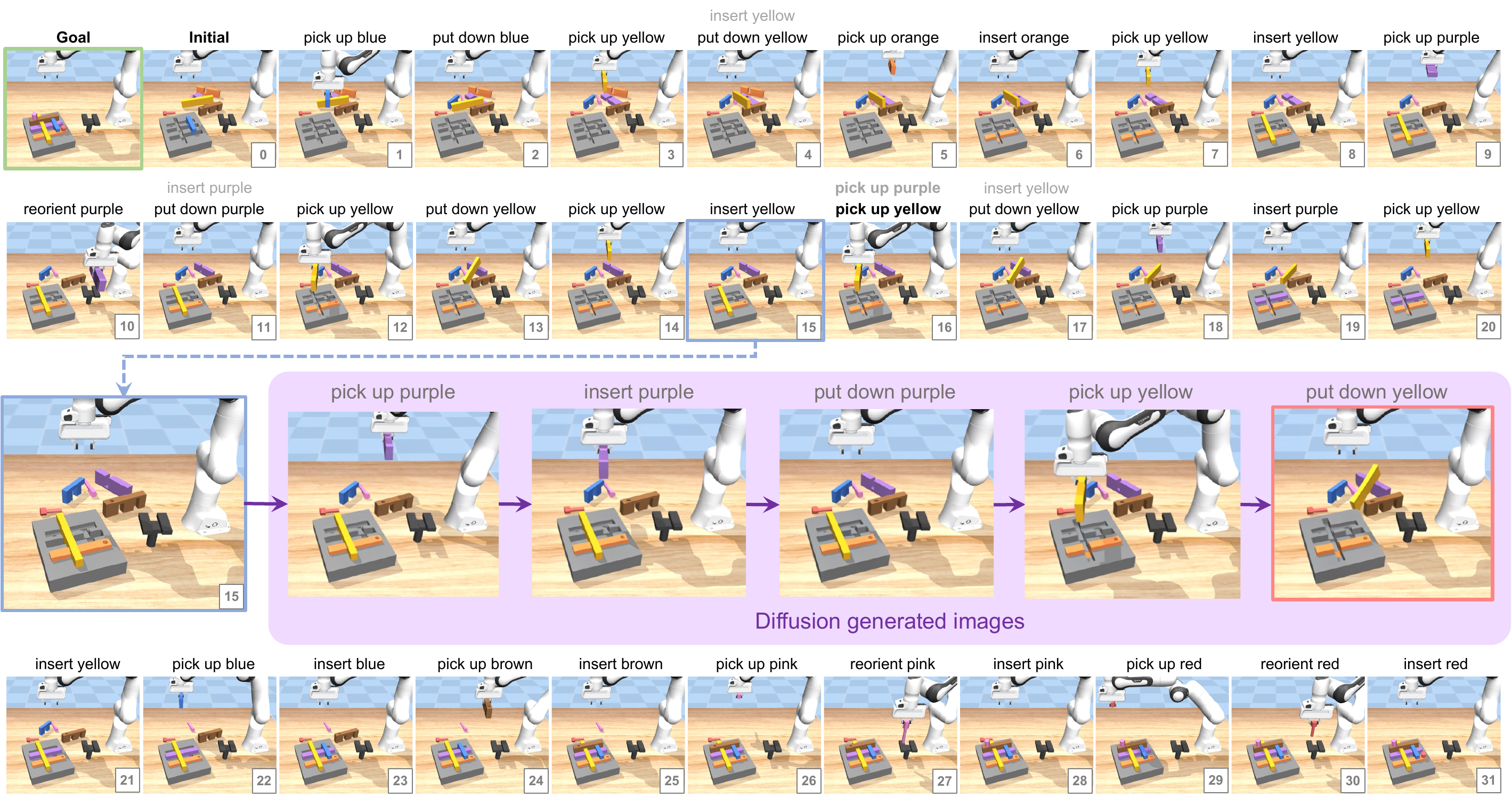}
    \caption{\textbf{Filmstrip of our method solving a complicated assembly task.} Frames are indexed by timestep. The goal image is in the top-left corner (with a green border). Each frame is the observation after executing the action (in black) above it. The other action in gray is the original action proposed by the VLM if it is revised after reflection. We highlight the reflection process at timestep 15, where the VLM first proposes an action to pick up the purple brick, but after reflection, it chooses to pick up the yellow brick instead as the generated future state (red-bordered image) shows little progress towards the goal.}
    \label{fig:filmstrip}
\end{figure*}
\begin{figure}[h!]
    \centering
    \includegraphics[width=0.49\textwidth]{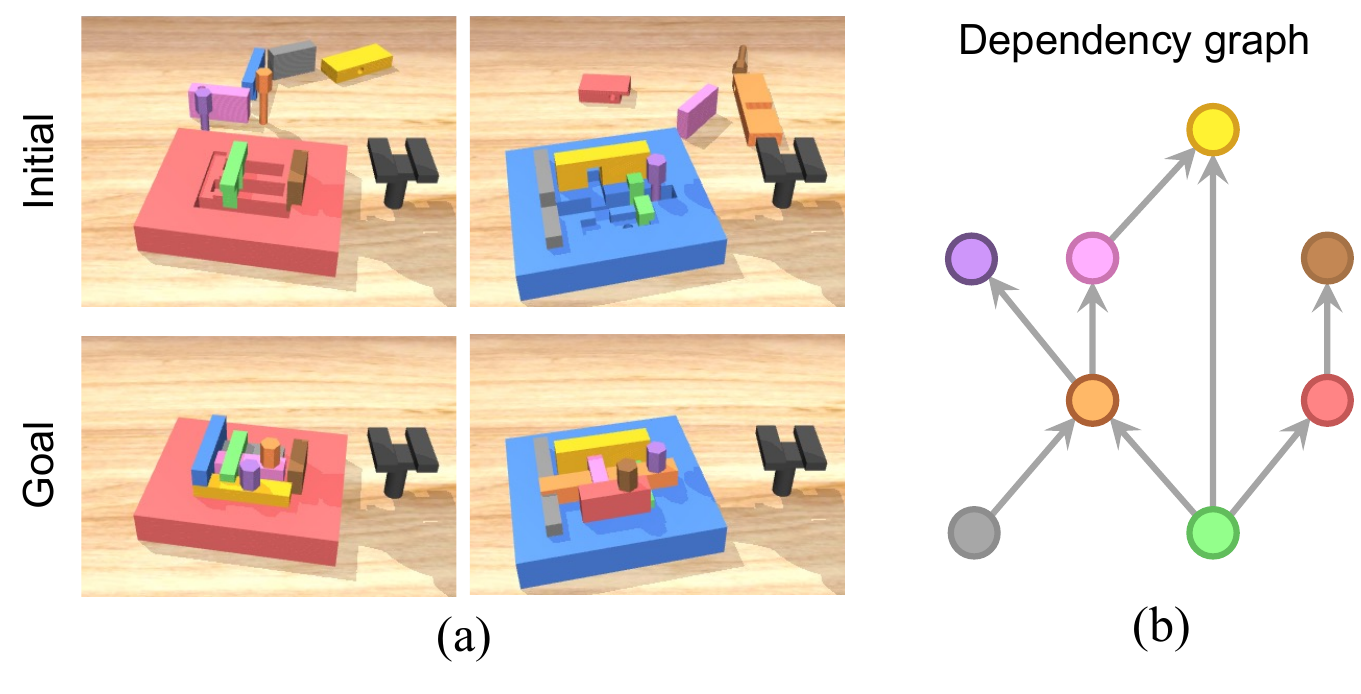}
    \vspace{-0.1in}
    \caption{\textbf{Task examples.} (a) Generated multi-stage manipulation tasks with interlocking pieces. Top: initial configurations. Bottom: goal configurations. See App.~\ref{sec:app_more_task_samples} for more examples. (b) The graph shows the dependencies between the objects in the blue assembly board on the left. Each node represents an object, and each directed edge indicates the predecessor object should be assembled before the successor object.}
    \label{fig:tasks}
\end{figure}

Inspired by~\citet{luo2024fmb}, we procedurally generated a suite of multi-stage long-horizon manipulation tasks that require understanding of physical interactions and reasoning about the effects of long-term action sequences. The task is initialized with a board and a set of small pieces randomly placed on a table. The goal is to fully assemble the board by inserting the pieces into the board one by one. Examples of the initial and goal configurations are shown in Fig.~\ref{fig:tasks}. Detailed task generation process is included in App.~\ref{sec:app_task_gen}. Notably, most tasks include inter-locking pieces so that they can be inserted into the board only in a specific order. This requires strategically choosing the object to be manipulated at each step and inferring possible interaction between this object and the other objects already in the board. 
As an example, Fig.~\ref{fig:tasks}(b) shows the dependencies between the pieces in one of the tasks. 
The interlocking feature further necessitates the agent’s ability to replan, enabling it to recover from failures caused by previous mistakes or bad initialization.

We focus on the high-level planning of this long-horizon manipulation task. We define a set of actions in the form of ``{\tt [act] [obj]}", where $\text{\tt [act]}\in \{\text{\tt pick up}, \text{\tt insert}, \text{\tt reorient}, \text{\tt put down}\}$ is an action primitive, and {\tt [obj]} denotes the object to be manipulated. Specifically, ``{\tt pick up}" grasps a piece that is not in hand and picks it up. It can then be inserted into the board using the ``{\tt insert}" action, or put back on the table using ``{\tt put down}". By invoking ``{\tt reorient}", the object in hand can be reoriented with the black fixture if necessary, so that it is in a suitable pose for insertion. Each action primitive is implemented as a rule-based script controller; however, integrating other low-level controllers, such as learning-based policies like behavior cloning, is also possible. We also designed an expert policy for the mentioned motor primitives, see App.~\ref{sec:app_expert} for implementation details.

\section{Experiments}\label{sec:experiment}
Our experiments evaluate the effectiveness of our method and analyze its key components. We aim to answer three key research questions. First, how well does our method perform in long-term planning, particularly when handling complex physical interactions? Second, how effectively does our method generalize across different object configurations and types, while maintaining the ability to reason and plan reactively in dynamic environments? Third, what is the impact of the reflection mechanism on the overall performance of our method?
To address these questions, we conduct comprehensive experiments comparing ReflectVLM against: (1) state-of-the-art VLM models tested in zero-shot fashions, (2) model-based planning approaches like MCTS, and (3) ablation studies examining the reflection mechanism. In this section, we first describe our experimental setup, followed by quantitative results and qualitative analysis.

\begin{figure*}[t!]
    \centering
    \includegraphics[width=0.99\textwidth]{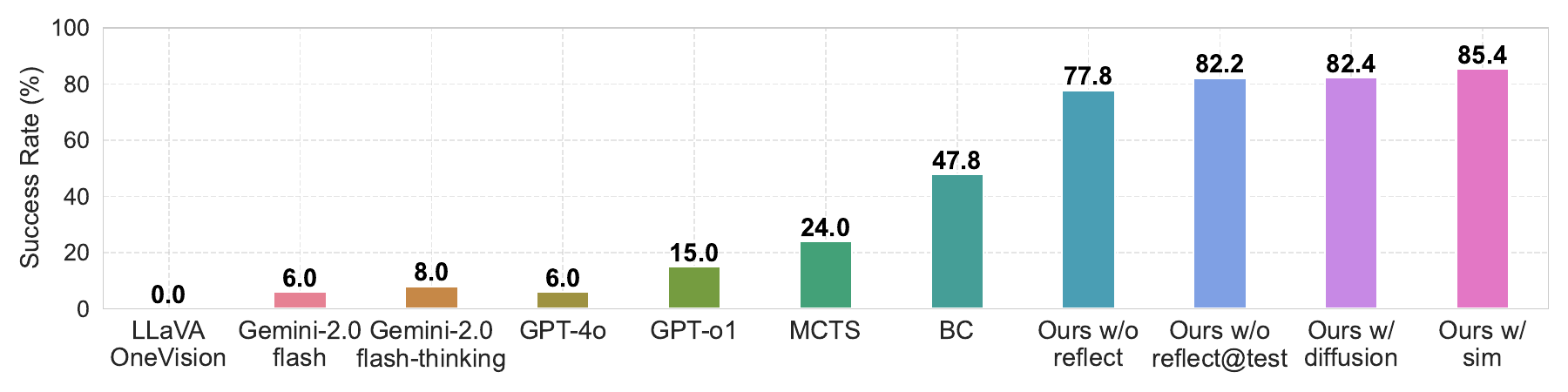}
    \vspace{-0.1in}
    \caption{\textbf{Performance of our method and baselines.} Success rate (\%) on 100 tasks. For the zero-shot test of state-of-the-art VLMs and MCTS, the experiments were conducted once; for other methods, the results are the average of five seeds.}
    \label{fig:main_comparison}
\end{figure*}

\subsection{Experiment Setup and Policy Training}
To evaluate the generalization capabilities of different models, we generate two distinct task sets: a training set using the procedure described in Sec.~\ref{sec:tasks}, and a separate evaluation set containing previously unseen configurations. The evaluation tasks are specifically designed to test generalization across varying object configurations, colors, and spatial arrangements. We particularly emphasize challenging scenarios that require sophisticated physical reasoning and multi-step planning. For instance, some tasks begin with objects in physically obstructing positions that prevent direct task completion - requiring the policy to first remove the obstructing pieces and then develop a new plan for the original objective.
Specifically, the training set contains 1000 different tasks, each generated task was randomized to five different initial spatial arrangements, these tasks are used to pre-train the VLM policy. At each iteration of post-training, we randomly sample 200 out of these 1000 tasks to further train the VLM policy with the reflection mechanism.
The evaluation set contains 100 different tasks that are unseen in the training set.

As mentioned in Sec.~\ref{sec:problem}, our method utilizes an oracle policy operating in the environment's symbolic state space to generate expert demonstrations for training. This oracle achieves a 97\% success rate across tasks, but importantly, it operates with access to ground-truth state information. In contrast, our VLM policy must rely solely on visual observations. While alternative data sources like human demonstrations could be used for training, we chose this oracle-based approach to systematically study our method's capabilities under controlled conditions.

During the policy pre-training phase, we utilize the oracle policy to provide action labels, then finetune an LLaVa-1.5-13B model~\cite{llava, llava1.5} with standard supervised learning loss. This pre-training used 5,000 expert demonstrations (1,000 unique tasks × 5 initial configurations per task). In the post-training phase, we use the same oracle policy to further train the VLM policy from the previous stage using the procedure described in Alg.~\ref{alg:training}. For each iteration of post-training, we collect 1k trajectories by rolling out the VLM policy in the environment to generate examples for fine-tuning. See App.~\ref{sec:app_training_details} for training details.

\subsection{Experiment Results}
In this subsection,  we report the results of different methods, and discuss their implications. Unless otherwise noted, numbers are reported across five runs, for some commercial VLMs such as GPT-o1, we only report one run due to cost consideration.

\begin{table}[t]
    \centering
        \centering
        \caption{{\textbf{Post-training performance} Success rates (\%) of post-training variants over the number of iterations.}}
        \label{tab:dagger_results}
        \begin{tabular}{l|ccc}
        \toprule Method             &  Iter. 1  & Iter. 2  & Iter. 3  \\
        \midrule
            w/o reflect	            &58.2    & 74.4 & \textbf{77.8}  \\
            w/o reflect@test	    &64.4    & 76.0 & \textbf{82.2} 	\\
            reflect w/ diffusion	&66.2    & 75.8 & \textbf{82.4} \\
            reflect w/ sim          &66.8   & 75.4 & \textbf{85.4} \\
        \bottomrule
        \end{tabular}
\end{table}
\begin{table}[t!]
    \centering
        \centering
        \caption{\textbf{Inference computation cost.} Inference wall clock time per step. MCTS result is averaged over 100 tasks and 1 seed; the others are averaged over 100 tasks and 5 seeds. All experiments are done on a single A100 GPU.}
        \label{tab:inference_cost}

        \begin{tabular}{l|r}
        \toprule Method &  Inference time (s) \\
        \midrule
            Ours w/o reflect@test & 0.45  \\
            Ours w/ diffusion	& 11.10 	\\
            Ours w/ sim & 6.05 \\
            MCTS & 391.42 \\
        \bottomrule
        \end{tabular}
\end{table}
\paragraph{VLM zero-shot} To evaluate the capabilities of state-of-the-art vision-language models, we tested several leading VLMs including LLaVAOneVision~\citep{llavaonevision}, Gemini-2.0-flash~\citep{google2024gemini}, Gemini-2.0-flash-thinking~\citep{google2024gemini}, GPT-4o~\citep{openai2024gpt4ocard}, and GPT-o1~\citep{openai2024openaio1card}. We specifically included Gemini-2.0-flash-thinking and GPT-o1 as they have demonstrated superior reasoning capabilities across various VLM benchmarks. As shown in Fig.~\ref{fig:main_comparison}, all models achieved notably low success rates on our tasks. Even GPT-o1, currently the most advanced proprietary model, succeeded in only 15 out of 100 tasks, primarily on simpler cases that did not require sophisticated physical reasoning about interlocking mechanisms. While Gemini-2.0-flash-thinking and GPT-o1 showed marginally better performance compared to other models, indicating some improved reasoning capabilities, their performance remains insufficient for solving our complex manipulation tasks. This significant performance gap confirms the necessity of our proposed method for handling physically-grounded reasoning tasks. Detailed evaluation procedures and results can be found in App.~\ref{sec:app_baseline_details}.

\paragraph{MCTS} To compare with model-based planning approaches, we implemented a VLM-based MCTS policy. This implementation uses our pretrained VLM policy as a base policy for generating candidate actions when expanding tree nodes, with value estimation provided by the oracle policy from the simulator. See App.~\ref{sec:app_baseline_details} for implementation details. As shown in Fig.~\ref{fig:main_comparison}, MCTS achieves a 24.0\% success rate—higher than zero-shot VLMs but lower than our method. Notably, while the pretrained VLM policy alone achieves a 47.8\% success rate, adding MCTS actually degrades performance. Our analysis revealed that although MCTS helped with some challenging tasks, it would sometimes incorrectly override valid plans from the base VLM policy. We found MCTS to be particularly challenging to tune effectively for our domain for several reasons: (1) it is highly sensitive to value function quality, (2) our tasks require nuanced physical reasoning that is difficult to capture in a value function, and (3) the possibility of succeeding from any state (by clearing the board and starting over) creates minimal value differences between states. These limitations highlight the advantages of our proposed method, which offers a lightweight, flexible approach that requires minimal tuning and can be readily integrated with any VLM policy.

\paragraph{ReflectVLM} Our full method outlined in Alg.~\ref{alg:training} and~\ref{alg:inference} incorporates reflection mechanisms in both training and inference phases. To systematically evaluate the impact of reflection, we conducted ablation experiments across several variants of our method.
As reported in Fig.~\ref{fig:main_comparison}, the variant without reflection in both training and inference achieved the lowest performance among our method's variants, though it still significantly outperformed the pretrained VLM baseline. The full method using a simulator during inference achieves the highest success rate, serving as an upper bound for our method's performance. When using a diffusion model instead of a simulator during inference, performance degrades slightly. This is unsurprising, as our tasks require nuanced understanding of physics and temporal dynamics—areas where current generative models still face challenges~\citep{kang2024farvideogenerationworld, motamed2025generativevideomodelslearn}. We expect our method's performance to improve as generative models advance.
We also report the post-training dynamics in Table~\ref{tab:dagger_results}. It's observed that the performance of all variants increases as more training is performed and the full method did achieve the highest performance as mentioned above.
While the absolute performance gap between variants may appear modest, the additional tasks solved by including reflection are qualitatively significant. These are typically complex scenarios requiring multiple replanning attempts, such as removing previously placed objects to explore alternative solutions—tasks the pretrained VLM consistently failed to solve. Notably, even without reflection during inference, our method achieves higher success rates than the pretrained baseline. This suggests that the natural language reflection prompts during training help the VLM policy develop better implicit reasoning capabilities. 
Fig.~\ref{fig:filmstrip} illustrates a representative example. In this complex task, the reflection mechanism iteratively revised suboptimal actions initially proposed by the VLM policy by identifying potentially unfavorable future states. This reflection capability proved crucial for success, as the long-horizon nature of the task required reactive planning and continuous adjustment of the solution strategy. 
Another point to consider is computation efficiency. Table~\ref{tab:inference_cost} shows the wall-clock time required per inference step. Compared to MCTS, our method requires only a fraction of the computation time while achieving substantially higher performance, making it particularly appealing as a lightweight and flexible solution for real-world applications.

\section{Discussion}\label{sec:discussion}
In this work, we presented a novel post-training strategy with reflection to improve VLM policies for long-horizon manipulation tasks, demonstrating superior planning capabilities with significantly less compute than traditional approaches like MCTS. Our current implementation opens up exciting future directions: while we currently use final outcomes for reflection due to VLM context constraints, future architectures with expanded context windows could enable richer intermediate feedback for more precise action refinement; the diffusion model's generation capabilities could be augmented with physical constraints and improved architectures to enhance prediction stability over longer horizons; and our single-round reflection approach could be extended to multiple rounds for iterative refinement while maintaining computational efficiency. 
We believe our method would benefit from continued advances in VLMs and generative models, and we hope it could establish a new foundation with broad applicability to sequential decision-making domains requiring visual understanding, physical reasoning, and long-horizon planning.

\input

\bibliography{example_paper}
\bibliographystyle{icml2025}

\newpage
\appendix
\onecolumn
\section{Task generation}\label{sec:app_task_gen}

\begin{figure}[h!]
    \centering
    \includegraphics[width=0.99\textwidth]{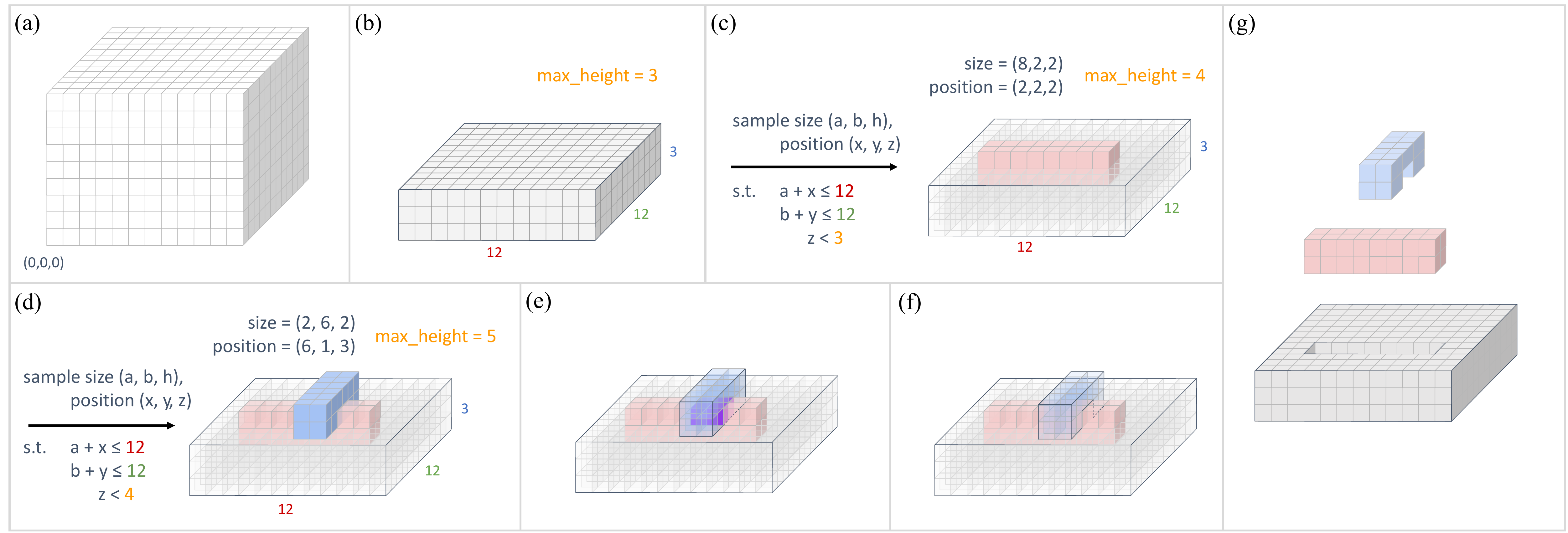}
    \caption{\textbf{Example of task generation.} (a) Voxel representation of the board. (b) Generating a base board. (c) Generating a red brick. (d) Generating another blue brick. (e) Critical voxels (highlighted in purple) at the intersection of the two bricks. (f) Handling intersection by assigning the critical voxels to the red brick. (g) Explosion view of the board consisting of three interlocking pieces.}
    \label{fig:task_gen}
\end{figure}

We here describe the procedure to generate assembly boards in detail with an example. A board is discretized into voxels and can be represented by a 3d array, where each value indicates the piece the voxel belongs to. Initially none of the voxels is occupied, so they are all set to an empty value 0, as shown in Fig.~\ref{fig:task_gen}(a). Then we iteratively add pieces to the board. We first sample the size of the base board, which is (12, 12, 3) in this example (Fig.~\ref{fig:task_gen}(b)). Then we set these voxels to 1 to indicate they belong to the base board. We also maintain a variable {\tt max\_height}, which represents the highest layer that contains non-zero voxels. To generate a brick, we sample its size and position subject to some constraints (Fig.~\ref{fig:task_gen}(c)). The first two constraints ensure that this brick is within the range of the base board, and the third constraint makes sure this brick will intersect with some previously generated brick. As before, we set the value of the red voxels to 2 to indicate they are from the new brick. Note that the voxels in the lower layer previously have a value of 1 since they belonged to the base board, but now their value is rewritten to 2. This also creates a hole on the base board. After generating this brick, we also update ${\tt max\_height}$ to 4 since we have 4 layers now. Fig.~\ref{fig:task_gen}(d) shows the process of generating another brick. As the new blue brick intersects with the old red brick at the four critical voxels highlighted in purple (Fig.~\ref{fig:task_gen}(e)), we can assign the value of these critical voxels to either that of the red one or the blue one. For example, keep these voxels to the red brick results in an opening on the blue one (Fig.~\ref{fig:task_gen}(f)). Stopping the generation process here gives us a board with three interlocking pieces, as shown in Fig.~\ref{fig:task_gen}(g).

\section{Samples of generated tasks}\label{sec:app_more_task_samples}
\begin{figure*}[h!]
    \centering
    \includegraphics[width=0.99\textwidth]{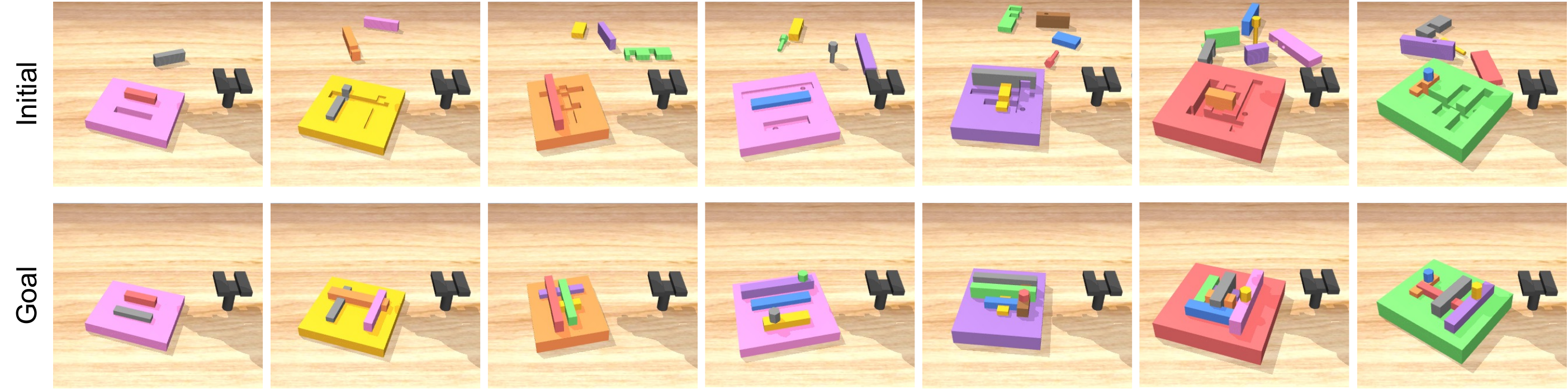}
    \caption{\textbf{Samples of generated tasks.} We procedurally generate a variety of multi-stage manipulation tasks, ranging from simple peg insertion to complex assembly tasks that contains multiple interlocking pieces. Top: initial configurations. Bottom: goal configurations.}
    \label{fig:more_task_samples}
\end{figure*}

\newpage
\section{Expert policy}\label{sec:app_expert}

The expert policy assumes access to the states of the objects in the simulator, such as the position and orientation of each piece. It is also provided with the dependency graph of the task, as discussed in Sec.~\ref{sec:tasks}. We define the status of each piece to be one of the following:
\vspace{-10pt}
\begin{itemize}
    \setlength{\topsep}{0pt}     
    \setlength{\partopsep}{0pt}  
    \setlength{\itemsep}{0pt}    
    \setlength{\parskip}{0pt}
    \item {\tt DONE}: if it is properly inserted into board;
    \item {\tt READY}: if it is not inserted yet but ready to be manipulated; 
    \item {\tt BAD\_B}: if it is in {\em{bad}} state since it is {\em{blocking}} other bricks, implying it needs to be removed;
    \item {\tt BAD\_D}: if it is in {\em{bad}} state since it is {\em{down}}, implying it needs to be reoriented;
    \item {\tt BLOCKED\_P}: if it is {\em{blocked}} since some {\em{predecessor}} brick(s) should be inserted before;
    \item {\tt BLOCKED\_S}: if it is {\em{blocked}} since some {\em{successor}} brick(s) is inserted before.
\end{itemize}
\vspace{-10pt}
Based on the status of each piece, we can also define a set of possible statuses for the whole assembly task:
\vspace{-10pt}
\begin{itemize}
    \setlength{\topsep}{0pt}     
    \setlength{\partopsep}{0pt}  
    \setlength{\itemsep}{0pt}    
    \setlength{\parskip}{0pt}
    \item {\tt DONE}: if the board is fully assembled, i.e., all pieces are in {\tt DONE} state;
    \item {\tt READY}: if some brick is in {\tt READY} or {\tt BAD\_D} state;
    \item {\tt BAD\_B}: if we need to reset some brick(s) to proceed as it is blocking other bricks.
\end{itemize}
\vspace{-10pt}
When queried, the expert policy first checks the status of each piece according to the simulation states, and decide the status of the whole task based on the statuses of all pieces. Then it decides the action to take following Algorithm~\ref{alg:expert}.

\begin{algorithm}
    \caption{Expert Policy}
    \label{alg:expert}
    \begin{algorithmic}[1]
    \REQUIRE {task status $status_\text{global}$, object in hand $obj_\text{hand}$, }

    \IF {$obj_\text{hand}$ is not None}
        \IF {all predecessors of $obj_\text{hand}$ are {\tt DONE}}
            \IF {$obj_\text{hand}$ is in {\tt BAD\_D} state}
                \STATE {\textbf{return} ``reorient $obj_\text{hand}$"}
            \ELSIF {$obj_\text{hand}$ is in {\tt BLOCKED\_S} state}
                \STATE {\textbf{return} ``put down $obj_\text{hand}$"}
            \ELSE
                \STATE {\textbf{return} ``insert $obj_\text{hand}$"}
            \ENDIF
        \ELSE
            \STATE {\textbf{return} ``put down $obj_\text{hand}$"}
        \ENDIF
    \ELSE
        \IF {$status_\text{global}$ == {\tt READY}}
            \STATE {choose an object $obj$ in {\tt READY} or {\tt BAD\_D} state}
            \STATE {\textbf{return} ``pick up $obj$"}
        \ELSIF {$status_\text{global}$ == {\tt BAD\_B}}
            \STATE {choose an object $obj$ in {\tt BAD\_B} state}
            \STATE {\textbf{return} ``pick up $obj$"}
        \ELSE
            \STATE {\textbf{return} ``done"}
        \ENDIF
    \ENDIF
    \end{algorithmic}
    \end{algorithm}

\section{Training details}\label{sec:app_training_details}
\subsection{VLM Policy}
\label{subsec:vlm_policy}

\begin{figure*}[t]
    \centering
    \includegraphics[width=0.95\textwidth]{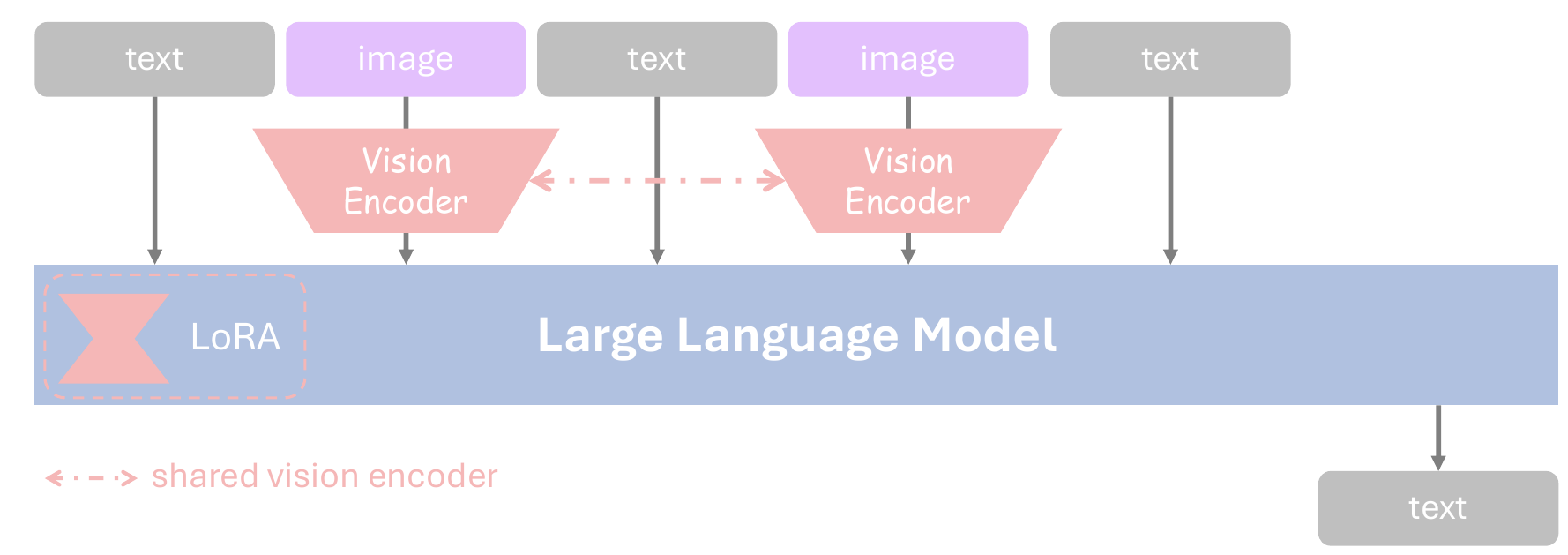}
    \caption{\textbf{Architecture of our VLM.} The model consists of a vision encoder and an LLM. We also add Low-Rank Adaptation (LoRA)~\cite{hu2022lora} layers to LLM for efficient adaptation. The input sequence contains interleaved images and text, where images are encoded into latent embeddings with a shared vision encoder. Finally, the concatenation of text and image embeddings are fed into VLM for multimodal reasoning.}
    \label{fig:vlm_arch}
\end{figure*}

\noindent \textbf{Architecture.} As shown in Fig.~\ref{fig:vlm_arch}, the architecture of our VLM consists of a vision encoder and a Large Language Model (LLM). By default, we use \texttt{clip-vit-large-patch14-336}~\footnote{\url{https://huggingface.co/openai/clip-vit-large-patch14-336}} as the vision encoder, and \texttt{vicuna-13b-v1.5}~\footnote{\url{https://huggingface.co/lmsys/vicuna-13b-v1.5}} as the LLM. We initialize our VLM with LLaVA-v1.5 weights~\footnote{\url{https://huggingface.co/liuhaotian/llava-v1.5-13b}} that are pre-trained on general visual instruction tuning datasets. Since our task prompts consist of interleaved images and text (refer to Sec.~\ref{sec:app_prompts}), we use a shared vision encoder to extract latent embeddings and concatenate them back to an input sequence.

\noindent \textbf{Training Parameters.} The full training parameters are listed in Table~\ref{tab:vlm_training_params}. For efficient adaptation of VLM to our task, we only finetune newly added LoRA~\cite{hu2022lora} layers. The rank of LoRA layers is 128 by default.

\begin{table}[h]
\caption{\textbf{Training parameters of VLM.}}
\label{tab:vlm_training_params}
\centering
\begin{tabular}{clcccccccc}
\toprule
\multirow{2}{*}{Res} & LoRA & Training & Batch & \multirow{2}{*}{Optimizer} & Warmup & \multicolumn{2}{c}{Learning rate}       & Weight & LR       \\
                     & Rank & Epoch    & Size  &                            & Epoch  & BC & Iter.~1,2,3 & Decay  & Schedule \\
                     \midrule
336px                & 128  & 1        & 128   & AdamW                      & 0.03   & 5e-5   & 1e-5                & 0.0    & Cosine  \\
\bottomrule
\end{tabular}
\end{table}

\subsection{Diffusion Dynamics Model}\label{app:ddm}

\noindent \textbf{Data Generation.} We generate 10$K$ different boards and use sub-optimal policies to collect transitions. The sub-optimal policies are implemented by setting a probability $p=\{0.2, 0.5, 0.7, 0.9, 1.0\}$ to replace the expert action by a random action. We collect 50$K$ trajectories; each has a maximum length of 50 and is terminated upon success. In total, we have about 1$M$ transitions. We randomly sample 50$K$  transitions for evaluation, and the rest is used for training.

\noindent \textbf{Training Parameters.} The full training parameters are listed in Table~\ref{tab:diffusion_training_params}. We initialize the Diffusion Dynamics Model with pretrained Instructpix2pix~\cite{brooks2022instructpix2pix}~\footnote{\url{https://huggingface.co/timbrooks/instruct-pix2pix}}.

\begin{table}[h]
\caption{\textbf{Training parameters of Diffusion Dynamics Models.}}
\label{tab:diffusion_training_params}
\centering
\begin{tabular}{c|cccccccccc}
\toprule
\multirow{2}{*}{Model} & \multirow{2}{*}{Res} & Training & Batch & \multirow{2}{*}{Optimizer} & Warmup & Learning & Weight & Beta1,     & Grad & LR       \\
                       &                      & Steps    & Size  &                            & Steps  & Rate     & Decay  & Beta2      & Norm & Schedule \\
                       \midrule
UNet         & 512px                & 20K      & 640   & AdamW                      & 2K     & 1e-4     & 0.01   & 0.9, 0.999 & 1.0  & Cosine   \\
Decoder         & 512px                & 4K       & 160   & AdamW                      & 1K     & 1e-7     & 0.01   & 0.9, 0.999 & 1.0  & Cosine  
\\
\bottomrule
\end{tabular}
\end{table}

\section{Prompts}\label{sec:app_prompts}

\subsection{Action proposal prompt}
\label{subsec:prompt}

\begin{tcolorbox}[width=\textwidth, colback=gray!10, colframe=black, boxrule=0.8pt]
There is a puzzle consisting of a board and several pieces with different colors on the table. The goal is to assemble the puzzle with the robot arm. In each step, one of the following four actions can be taken: pick up [obj], put down [obj], reorient [obj], and insert [obj], where [obj] refers to the piece to be manipulataed. The image of the goal state is: \textless image\textgreater. The image of the current state is: \textless image\textgreater. The most recently executed actions are: {\tt\{history\}}. What action should be taken next? Note that [obj] should be a color chosen from the following list: {\tt\{colors\}}.
\end{tcolorbox}

\subsection{Reflection prompt}

\begin{tcolorbox}[width=\textwidth, colback=gray!10, colframe=black, boxrule=0.8pt]
There is a puzzle consisting of a board and several pieces with different colors on the table. The goal is to assemble the puzzle with the robot arm. In each step, one of the following four actions can be taken: pick up [obj], put down [obj], reorient [obj], and insert [obj], where [obj] refers to the piece to be manipulataed. The image of the goal state is: \textless image\textgreater. The image of the current state is: \textless image\textgreater. The most recently executed actions are: {\tt\{history\}}. The next five steps planned by the model is {\tt\{init\_plan\}}, from which we are going to only execute the first action. Note that if the full plan was executed sequentially, the future state would be: \textless image\textgreater. What action should be taken for the immediate next step? Note that [obj] should be a color chosen from the following list: {\tt\{colors\}}. You can modify the initial plan if it leads to an undesired future state.
\end{tcolorbox}

\section{Baseline details}\label{sec:app_baseline_details}
\subsection{Zero-shot VLMs}
We prompt state-of-the-art close-sourced and open-sourced VLMs for zero-shot evaluation, including LLaVA-Onevision, Gemini-2.0 (\texttt{gemini-2.0-flash-exp}), Gemini-2.0-thinking (\texttt{gemini-2.0-flash-thinking-exp-1219}), GPT-4o and GPT-o1. We resize all input images to 336$\times$336 pixels for fair comparisons with our model. We set the generation temperature and max planing step to 0 and 50. The evaluation prompt is:
\begin{tcolorbox}[width=\textwidth, colback=gray!10, colframe=black, boxrule=0.8pt]
You are an intelligent robot equipped with cameras and robotic arms, your primary task is to observe and interact with the objects on the desktop.
\\
\\
\{Action proposal prompt (Sec.~\ref{subsec:prompt})\}
\\
\\
You can only output the action, e.g., pick up red. Do not output anything else.
\end{tcolorbox}

\begin{figure}
    \centering
    \includegraphics[width=\linewidth]{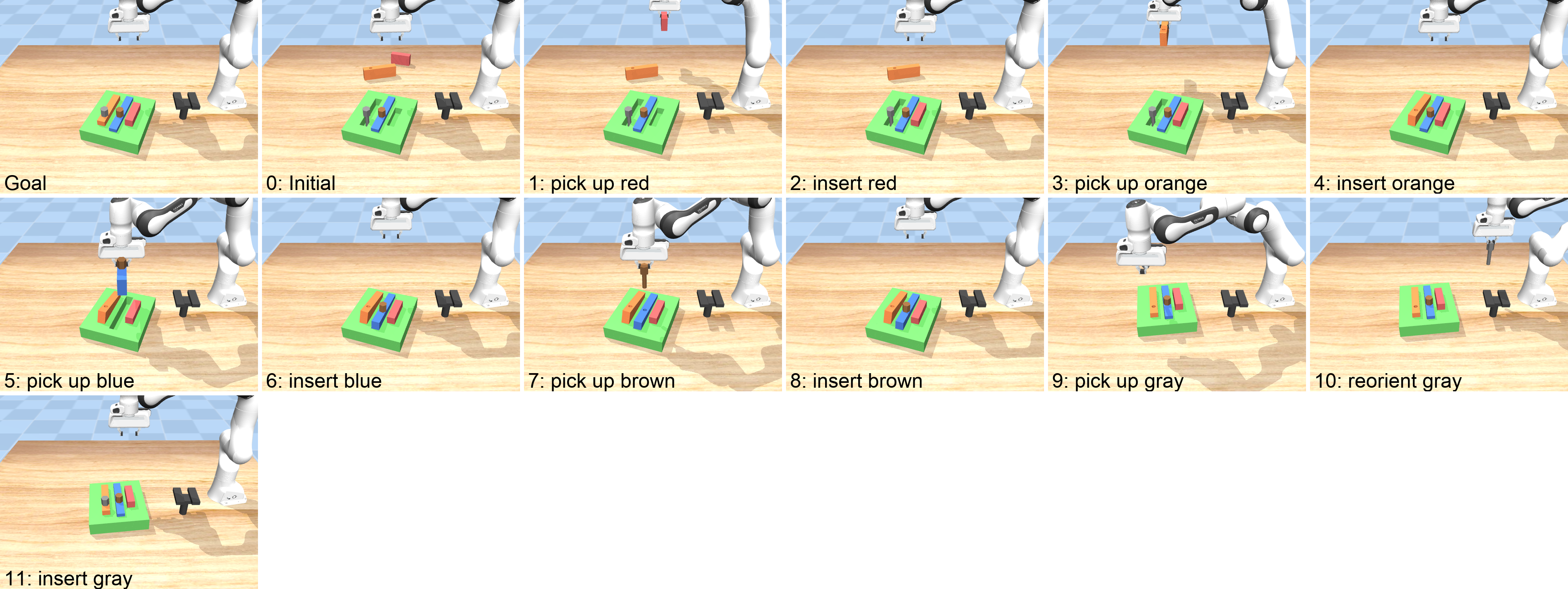}
    \\
    \vspace{5mm}
    \includegraphics[width=\linewidth]{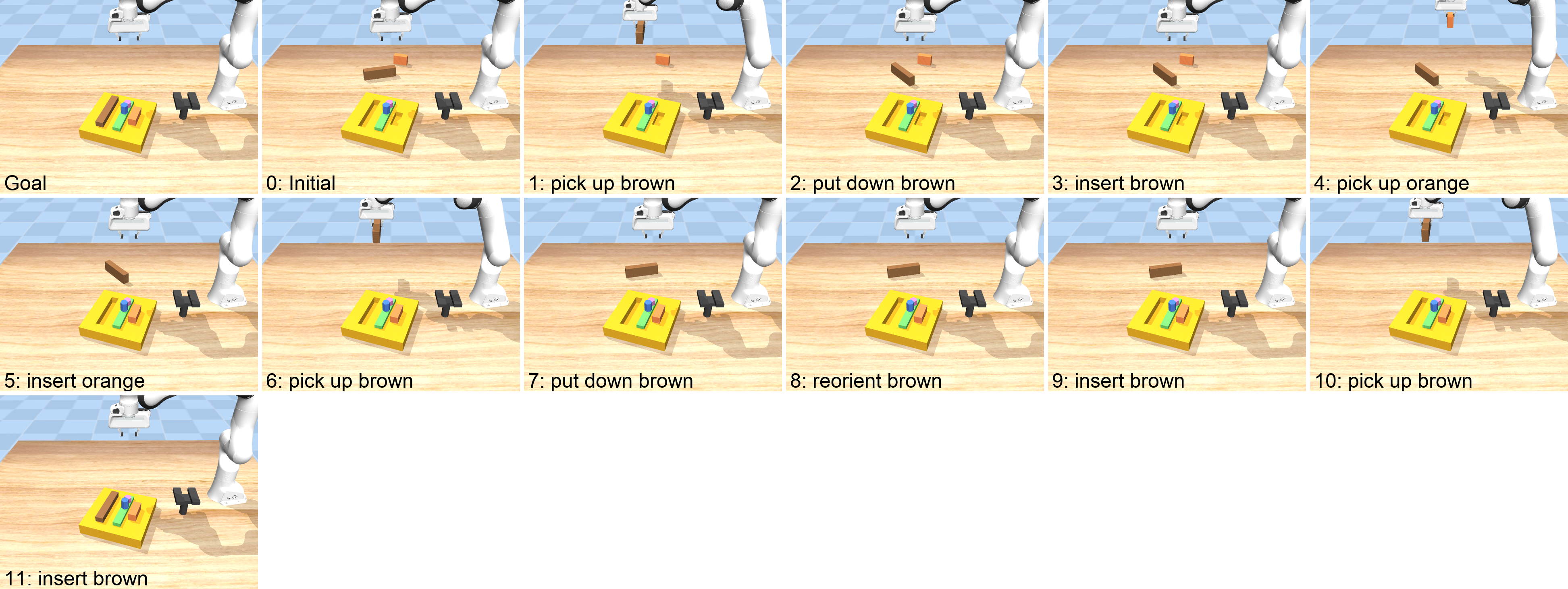}
    \\
    \vspace{5mm}
    \includegraphics[width=\linewidth]{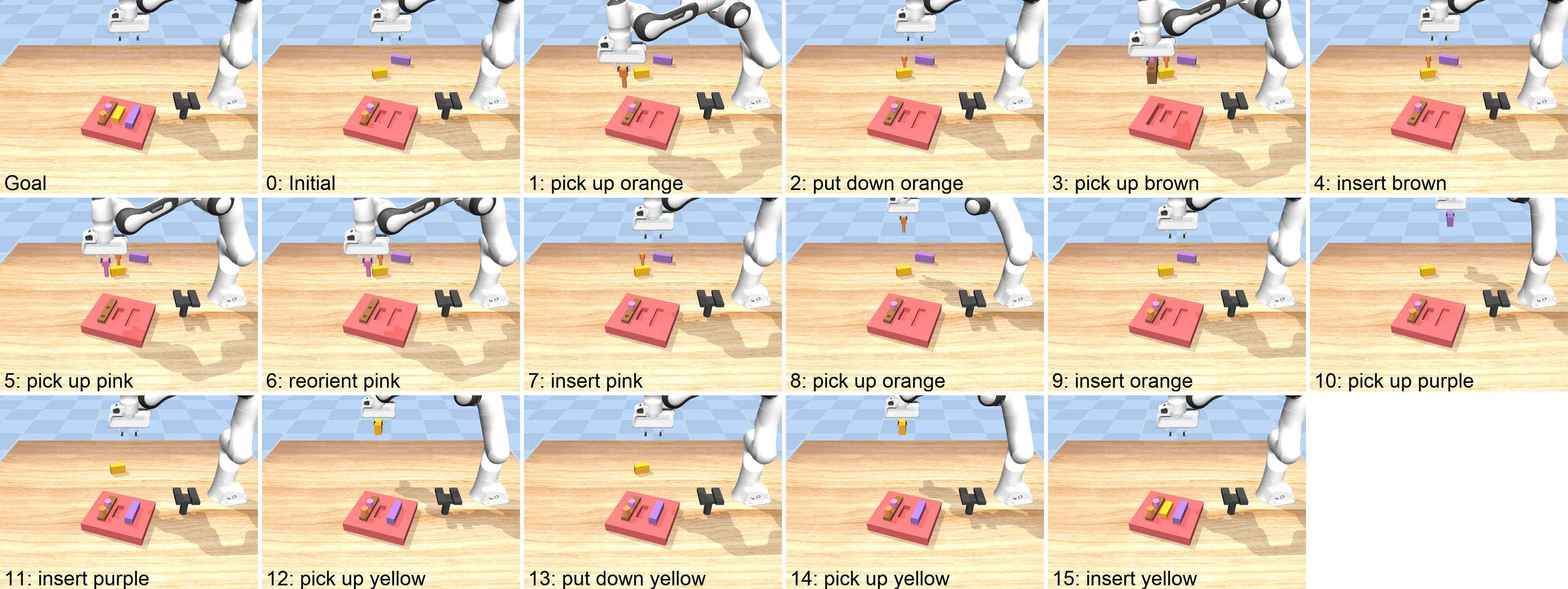}
    \caption{\textbf{Success cases of zero-shot VLMs.} Top: Gemini-2.0; Middle: Gemini-2.0-Thinking; Bottom: GPT-4o.}
    \label{fig:zero_shot_success_part1}
\end{figure}

\begin{figure}
    \centering
    \includegraphics[width=\linewidth]{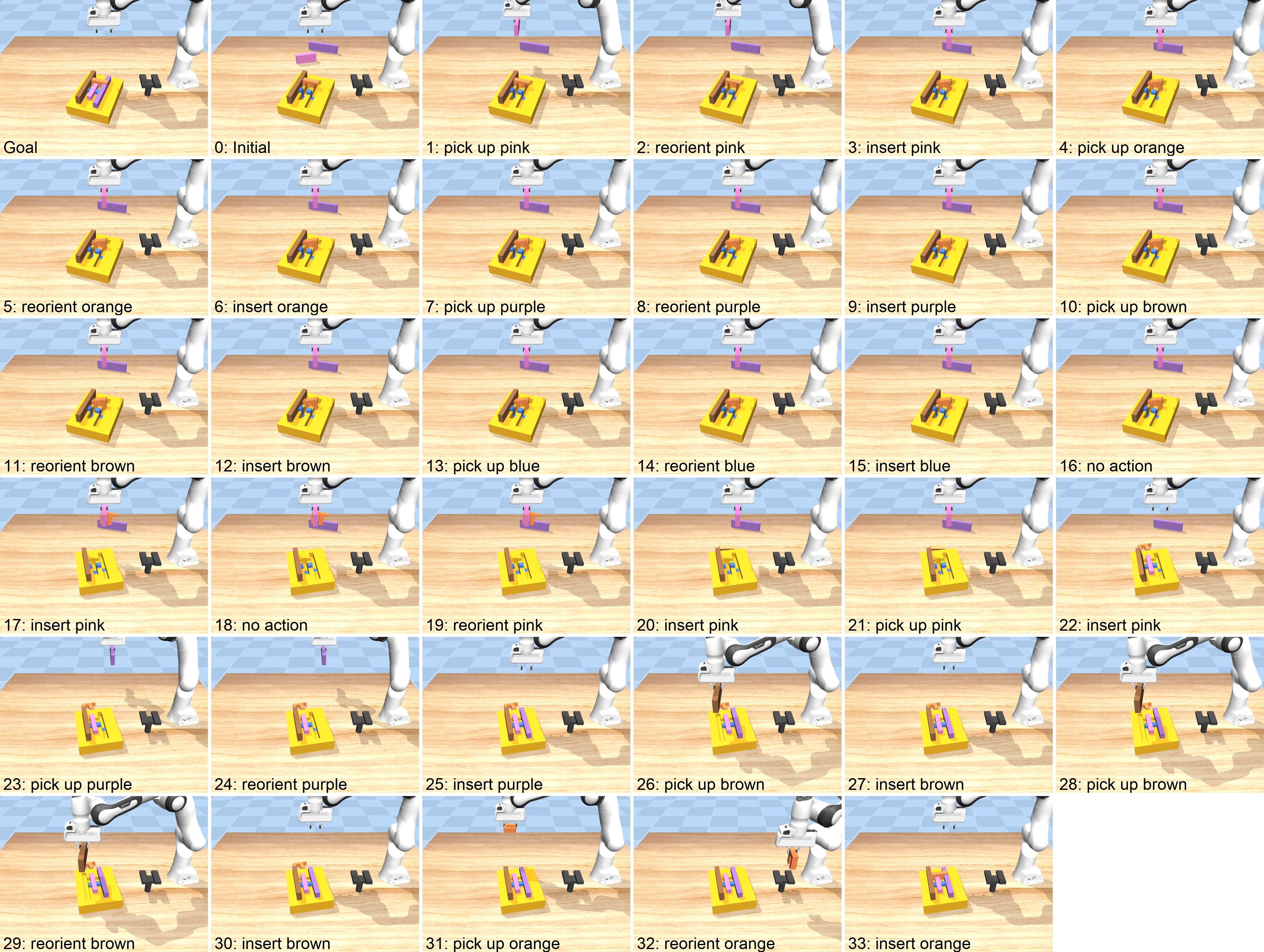}
    \caption{\textbf{Success cases of zero-shot VLMs (GPT-o1).}}
    \label{fig:zero_shot_success_part2}
\end{figure}

\begin{figure}[htp]
    \centering
    \includegraphics[width=\linewidth]{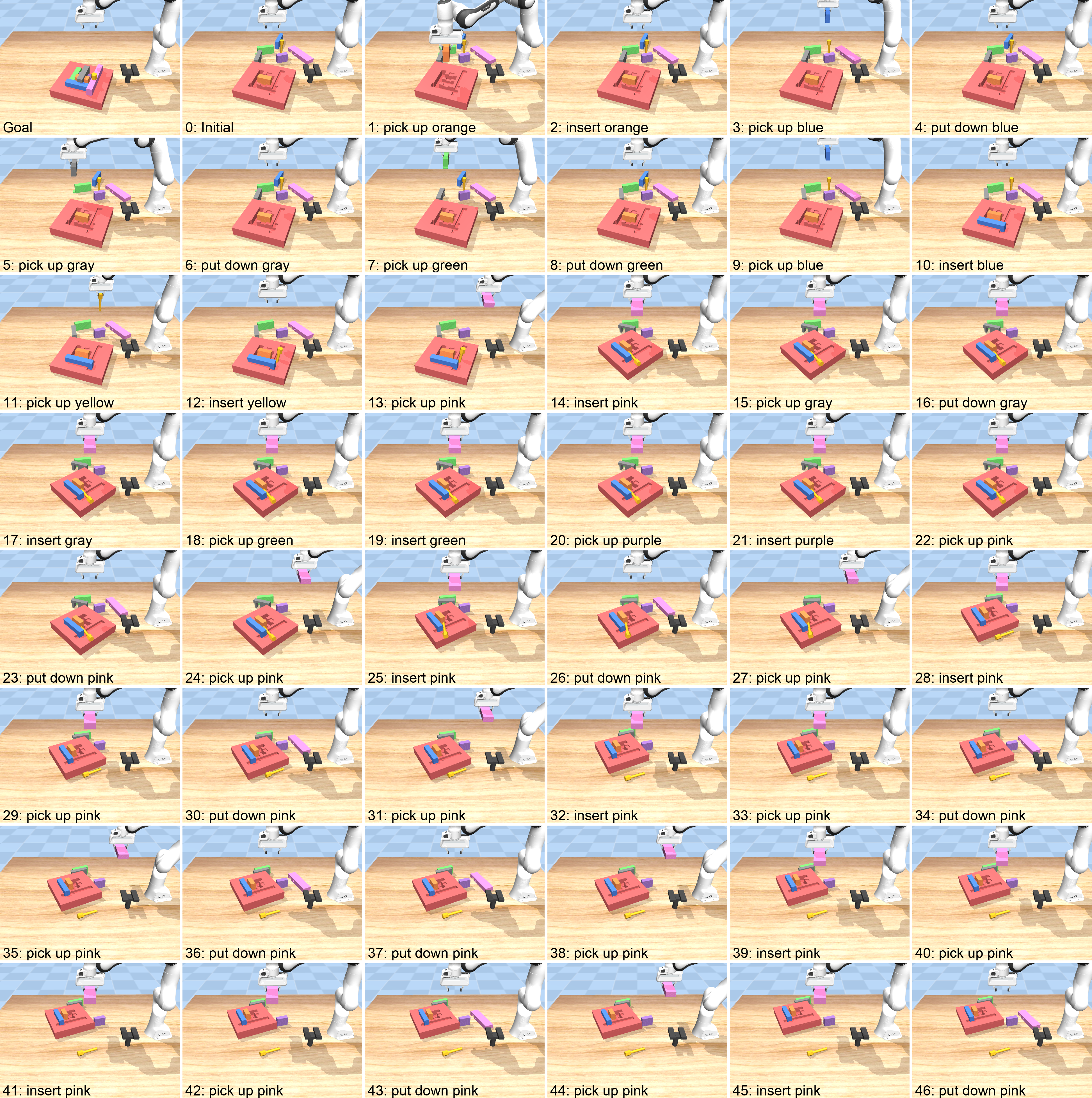}
    \caption{\textbf{Failure case of Gemini-2.0.}}
    \label{fig:failure_gemini_2}
\end{figure}

\begin{figure}[htp]
    \centering
    \includegraphics[width=\linewidth]{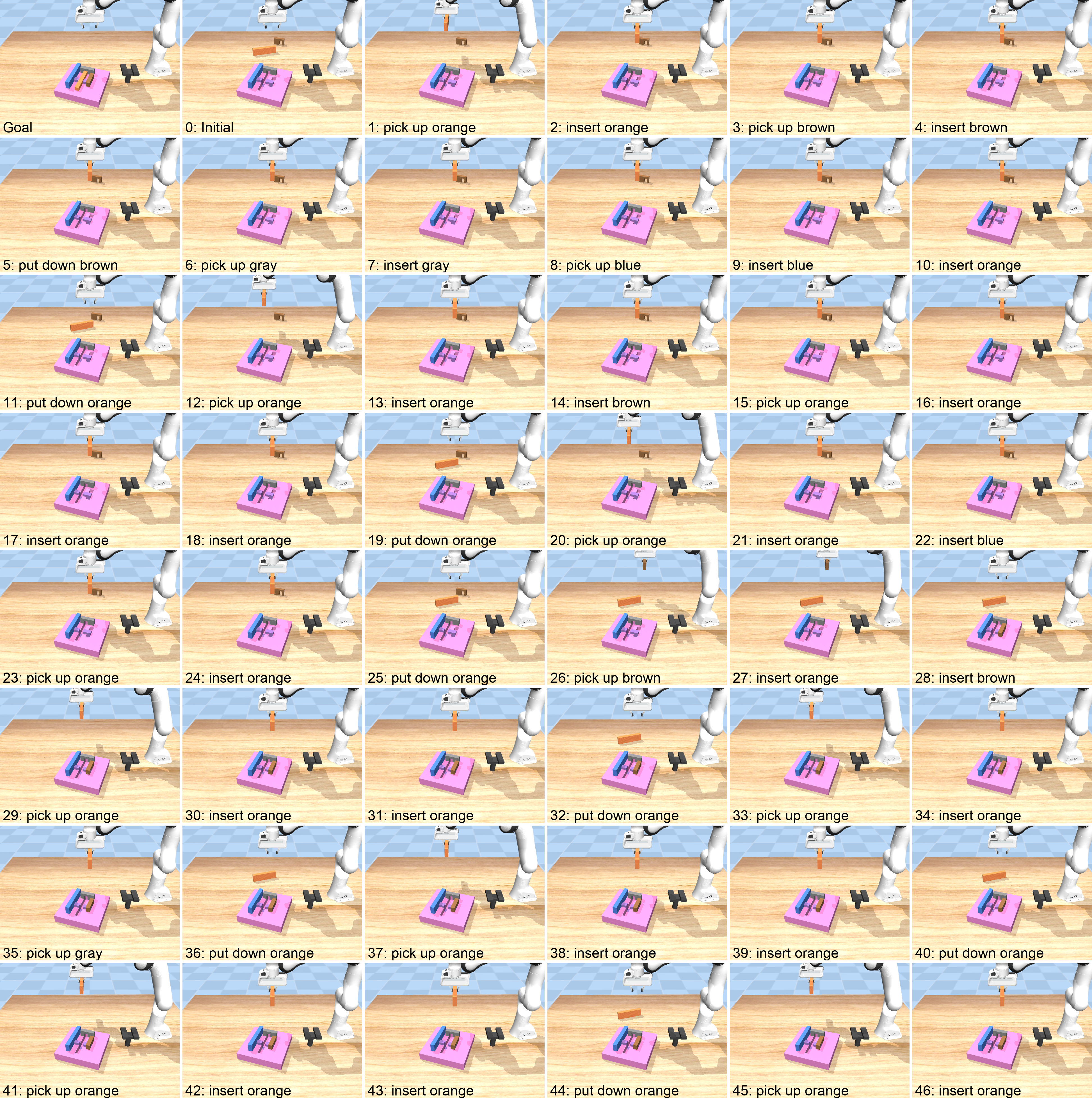}
    \caption{\textbf{Failure case of Gemini-2.0-Thinking.}}
    \label{fig:failure_gemini_2_think}
\end{figure}

\begin{figure}[htp]
    \centering
    \includegraphics[width=\linewidth]{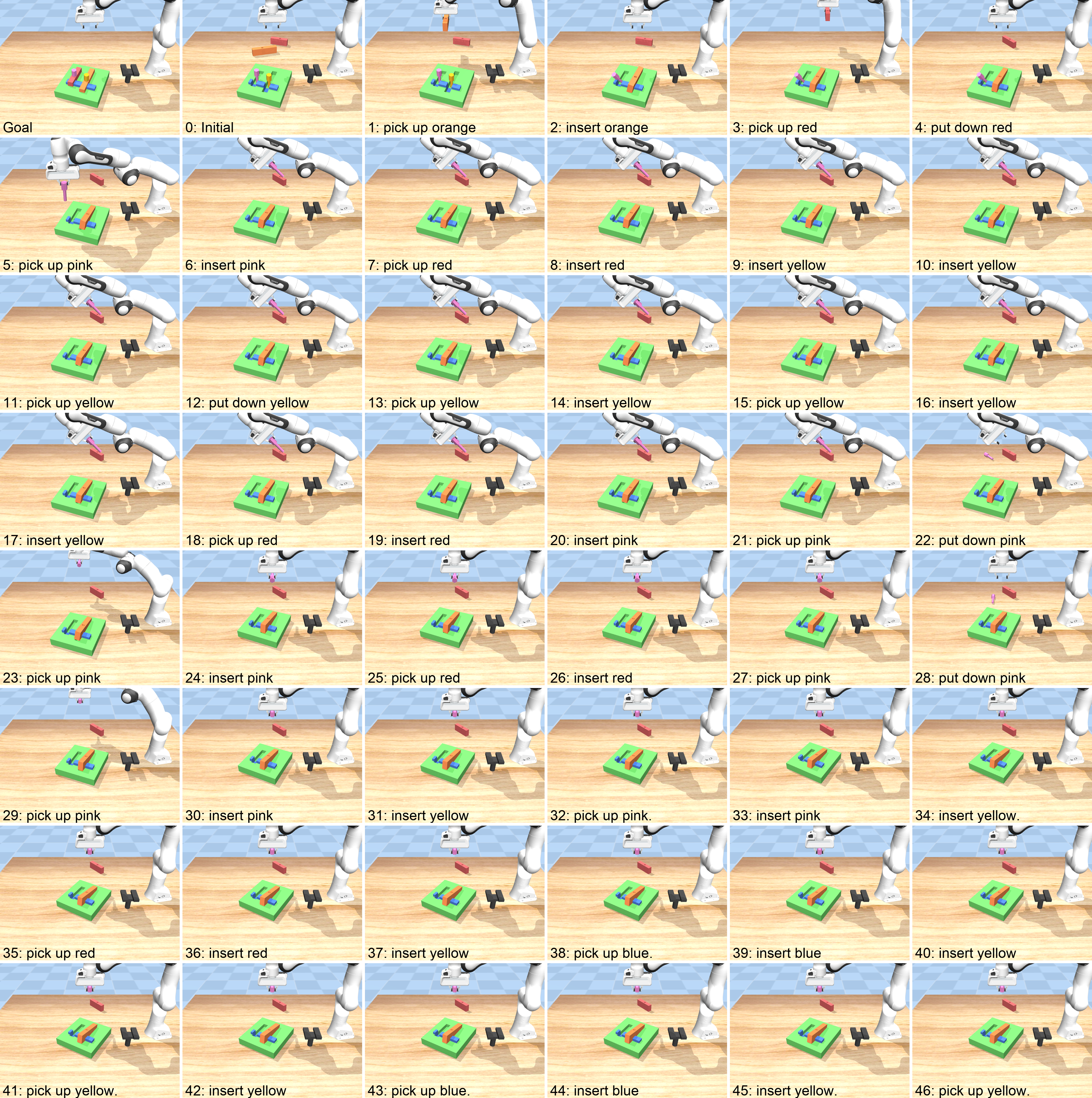}
    \caption{\textbf{Failure case of GPT-4o.}}
    \label{fig:failure_gpt4o}
\end{figure}

\begin{figure}[htp]
    \centering
    \includegraphics[width=\linewidth]{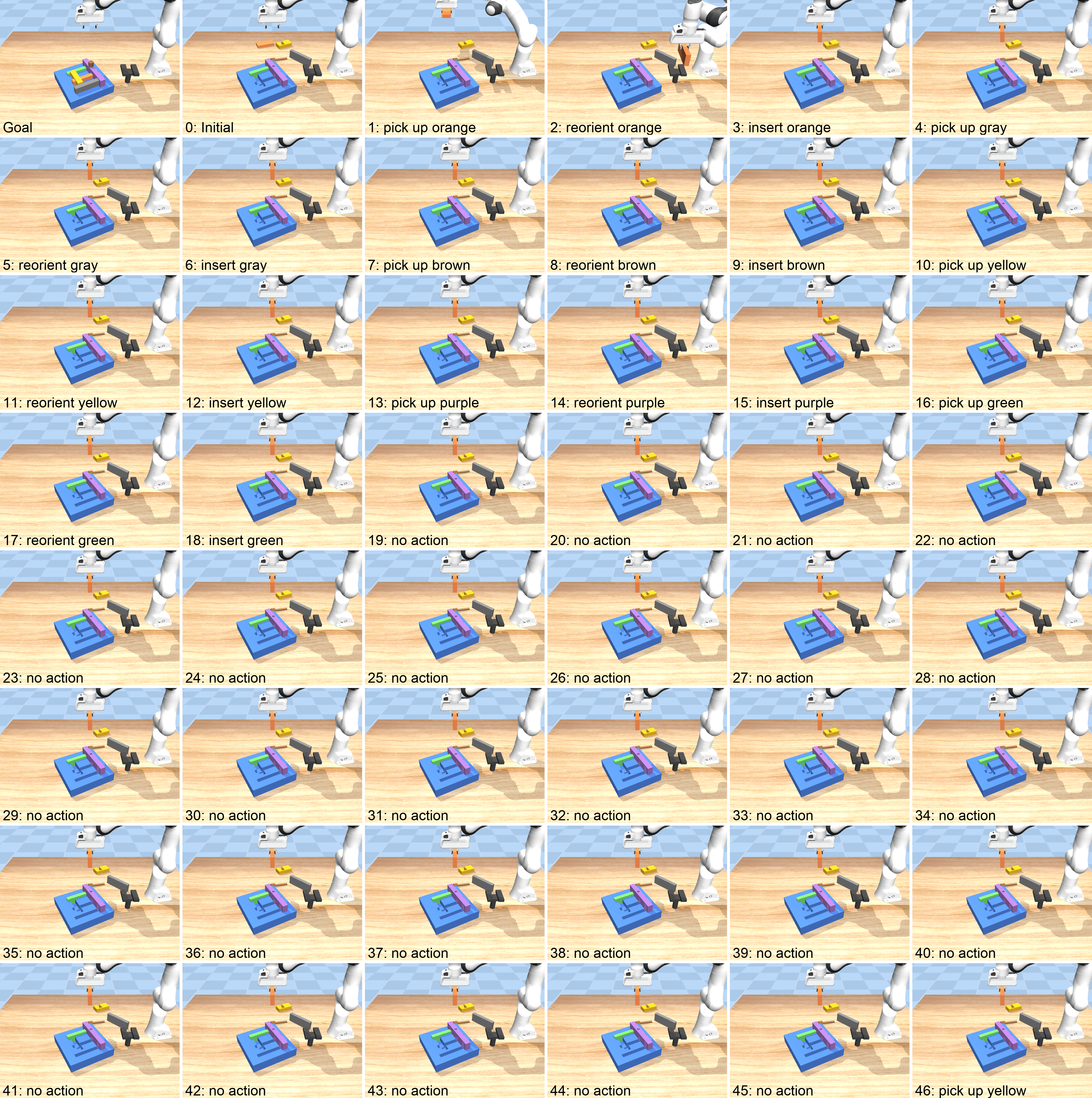}
    \caption{\textbf{Failure case of GPT-o1.}}
    \label{fig:failure_gpto1}
\end{figure}

Since the instruction following capability of LLaVA-Onevision is quite limited, we cannot extract valid actions from its response. For other close-sourced VLMs, we list the detailed evaluation results in Table~\ref{tab:zero_shot_details}. We also visualize some success cases in \cref{fig:zero_shot_success_part1,fig:zero_shot_success_part2}, and failure cases in \cref{fig:failure_gemini_2,fig:failure_gemini_2_think,fig:failure_gpt4o,fig:failure_gpto1}.

\begin{table}[thb]
\caption{\textbf{Detailed evaluation results of zero-shot VLMs.}}
\label{tab:zero_shot_details}
\centering
\begin{tabular}{llccc}
\toprule
Model                   & Success Trajectory ID / Planing Steps                & Max Steps           & Min Steps          & Avg Steps             
\\
\toprule
Gemini-2.0              & 5/6, 12/4, 16/18, 47/11, 60/4, 86/6                  & 18                  & 4                  & 8.2                   \\
\midrule
Gemini-2.0-Thinking     & 5/6, 12/4, 40/20, 47/16, 50/8, 60/8, 86/10, 90/11    & 20                  & 4                  & 10.4                  \\
\midrule
GPT-4o                  & 12/15, 16/5, 19/4, 47/10, 60/4, 90/6                 & 15                  & 4                  & 7.3                   \\
\midrule
\multirow{2}{*}{GPT-o1} & 12/9, 16/6, 17/15, 47/8, 50/16, 58/18, 60/14, 62/33, & \multirow{2}{*}{33} & \multirow{2}{*}{4} & \multirow{2}{*}{13.1} \\
                        & 66/6, 67/12, 72/32, 77/9, 85/9, 86/6, 90/4           &                     &                    &
                        \\ \bottomrule
\end{tabular}
\end{table}

\subsection{MCTS}

We implemented MCTS similar to AlphaGo Zero~\cite{alphagozero} but with a VLM policy for action proposal and a heuristic value estimator. States and actions are represented by nodes and edges, respectively. The algorithm iteratively expands the search tree and estimates the value for different actions. We store the visit count $N(s,a)$, total action value $W(s,a)$, and action value $Q(s,a)=W(s,a)/N(s,a)$ on edges. Each iteration consists of three phases: (1) select, (2) expand, and (3) backup.

In select phase, it traverses the tree by selecting the edge that has the largest action value $Q(s, a)$ plus an upper confidence bound $U(s, a)=c_\text{explore}{\sqrt{\sum_{a'} N(s,a')}}/{(1+N(s,a))}$, 
where $c_\text{explore}$ is the factor to balance exploring less visited edges and exploiting edges with high value. We use $c_\text{explore}=0.5$ in our experiments. If there is no actions associated to a node yet, it samples 5 top-likelihood actions with the VLM, with duplicates removed, and adds them to the node. 

In expand phase, it expands the selected edge by simulating the action in the simulator, getting the next state, and adding the new state to the tree as a new node. It then estimates the value of the new state by rolling out the expert policy from that state. The estimated value is $V=\exp(-\lambda S)$, where $S$ is the number of steps the expert policy takes to reach the goal from the new state, and $\lambda=0.1$ is a scaling factor.

In backup phase, it updates the statistics of the edges on the path from the root to the expanded node: $N(s,a)\leftarrow N(s,a)+1$, $W(s,a)\leftarrow W(s,a)+V$, and $Q(s,a)\leftarrow W(s,a)/N(s,a)$.

The search completes after 50 iterations. Among all actions connected to the root node, the action with the highest $Q$ value is chosen to execute. We replan with MCTS at each timestep. 

\begin{figure}[htp]
    \centering
    \includegraphics[width=\linewidth]{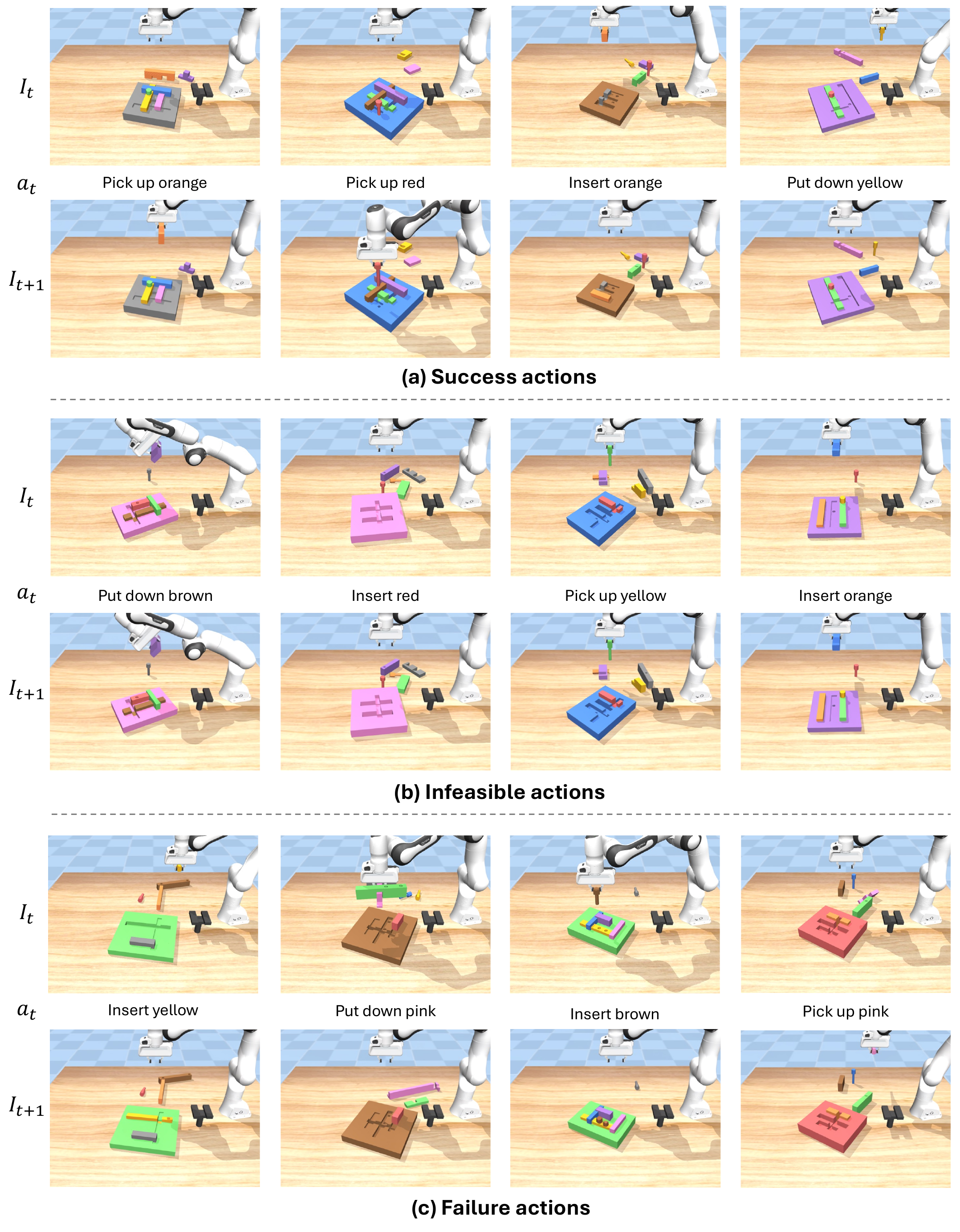}
    \caption{\textbf{Examples of Diffusion Dynamic Models.}}
    \label{fig:diffusion_cases}
\end{figure}

\end{document}